\documentclass[review,10pt]{JMtemplate}

\usepackage{lastpage}
\usepackage[utf8]{inputenc}
\usepackage[T1]{fontenc}

\usepackage{amsmath}
\usepackage{amsfonts}
\usepackage{booktabs}
\usepackage{multirow}
\usepackage{graphicx}
\usepackage{nicefrac}
\usepackage{microtype}
\usepackage{xcolor}
\usepackage{placeins}
\usepackage{lipsum}
\usepackage{longtable}
\usepackage{booktabs}      
\usepackage{threeparttable}
\usepackage{multicol}
\usepackage{array}
\usepackage{subcaption}
\usepackage{changepage}
\usepackage{pdflscape}
\usepackage{makecell}
\usepackage{adjustbox}
\usepackage{changepage}
\usepackage{caption}
\usepackage[figuresright]{rotating}
\graphicspath{{media/}}
\usepackage{marvosym}
\usepackage{algorithm}
\usepackage{algorithmic}

\usepackage{mdframed}

\usepackage[
    colorlinks=true,    
    linkcolor=blue,     
    urlcolor=blue,      
    citecolor=green,    
    filecolor=magenta   
]{hyperref}
\usepackage{textcomp}

\begin{document}
\begin{frontmatter}
\title{GRUET: Quantifying Uncertainty of Agentic Reasoning-and-Acting Processes}

\author{\textbf{Shuang Liang}\textsuperscript{\rm 1,2} \quad
\textbf{Xin-Yu Hu}\textsuperscript{\rm 1,2} \quad
\textbf{Shao-Qun Zhang}\textsuperscript{\rm 1,2,\Letter} \\[0.3em]
\small \textsuperscript{1} National Key Laboratory for Novel Software Technology, Nanjing University, China.\\
\small \textsuperscript{2} School of Intelligent Science and Technology, Nanjing University, China.\\
\small \texttt{ zhangsq@lamda.nju.edu.cn }
}

\begin{abstract}
Agents have attracted considerably increasing attention due to the power of executing both Reasoning and Acting (ReAct) in open and dynamic environments. The ReAct process typically exhibits a multi-turn trajectory in which one drives Large Language Models (LLMs) to generate both reasoning chains and task-specific actions in an interleaved manner. However, agents often suffer from significant uncertainty, where identical tasks yield divergent trajectories; trajectories with higher uncertainty often produce incomprehensible behaviors, severely undermining agent credibility. This work conjectures that such trajectory-level uncertainty frequently stems from cumulative turn-level reasoning uncertainty induced by LLMs; the latter often exhibits a collection of branches of divergent reasoning chains and their resulting actions. Built upon this, we present the Graph-based Reasoning UncErtainty in Trajectories (GRUET) method for the uncertainty quantification of ReAct, comprising turn-level reasoning uncertainty quantification and trajectory-level uncertainty aggregation; the former precisely quantifies reasoning uncertainty via modeling the reasoning space spanned by potential reasoning branches as a graph and then approximating the reasoning space complexity with graph complexity, while the latter employs simple aggregation strategies for quantifying the overall trajectory credibility. Empirical evaluations across nine LLMs and five benchmarks validate the effectiveness of our proposed GRUET in terms of selective generation performance, measured by AUROC, AUPRC, and AUARC.

\textit{Key words:} Large Language Models, Agents, ReAct, Trajectory-level Uncertainty Quantification, Turn-level Reasoning Uncertainty Quantification
\end{abstract}
\end{frontmatter}

\begin{figure*}[th]
    \centering
    \includegraphics[width=\linewidth]{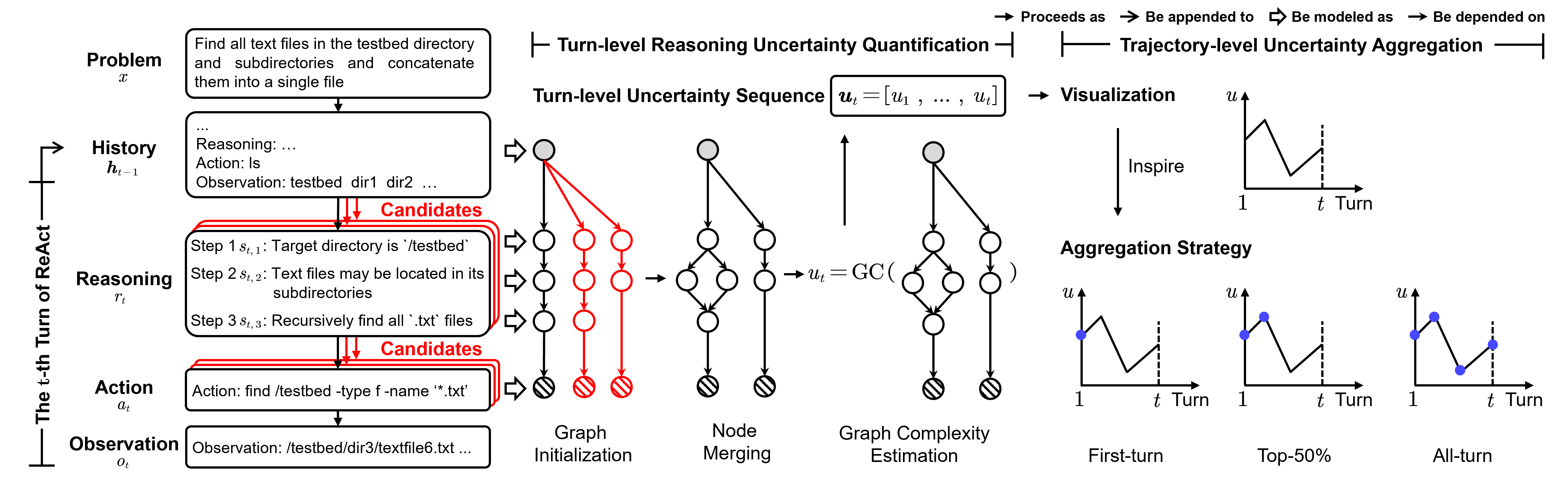}
    \caption{Workflow of our proposed GRUET.}
    \label{fig:overview_ureact}
\end{figure*}

\section{Introduction}  \label{sec:intro}
The recent emergence of agents has transformed Large Language Models (LLMs) from simple question-answering tools into autonomous workers through the multi-turn Reasoning and Acting (ReAct)~\cite{Yao2023react} paradigm, where the agent drives LLMs to generate both a reasoning chain and the resulting textual action for interacting with environments at each turn, with this procedure repeated to form a trajectory. However, when employing agents, developers frequently encounter uncertainty that manifests as a proliferation of divergent trajectories, even when tackling the same task, thereby severely compromising the credibility of agents~\cite{oh2026uncertainty,shapira2026agents}. Thus, it is necessary and significant to quantify the uncertainty of the ReAct process, while little attention has been paid to this topic.

This work focuses on the Uncertainty Quantification (UQ) of the ReAct process. We conjecture that such trajectory-level uncertainty frequently stems from cumulative LLM-induced turn-level uncertainty, which consists of turn-level reasoning uncertainty and turn-level acting uncertainty; the former often manifests as branches of divergent reasoning chains and their resulting textual actions, whereas the latter typically exhibits variability in executable actions such as tool-calling. Turn-level acting uncertainty is fundamentally encapsulated within turn-level reasoning uncertainty because the executable actions are deterministically parsed from the textual actions within the reasoning chains~\cite{yang2024sweagent}. Therefore, achieving precise quantification of turn-level reasoning uncertainty can effectively reflect turn-level uncertainty, which in turn contributes to accurately capturing the overall trajectory-level uncertainty.

There have been efforts on UQ methods for LLM reasoning, which can be roughly divided into three categories involving reflexive-based, information-based, and diversity-based methods~\cite{vashurin2025benchmarking}. Hence, it is intuitive to implement turn-level reasoning UQ by exploiting one of these methods and averaging the resulting uncertainty scores across turns to measure trajectory-level uncertainty. Unfortunately, such LLM UQ extensions usually suffer from limited performance in the downstream task of selective generation~\cite{ren2023out}. Specifically, a recent study~\cite{oh2026uncertainty} showed that extensions of reflexive-based UQ, such as Verbalized Confidence~\cite{xiong2024can}, and information-based UQ like Perplexity~\cite{fomicheva2020ppl}, typically exhibit limited selective generation performance comparable to that of a random uncertainty score generator. We evaluated this performance for diversity-based extensions, such as Semantic Entropy~\cite{farquhar2024detecting}, EigV~\cite{lin2024generating}, and SentSAR~\cite{duan2024sar}, as shown in Table~\ref{tab:uq_performance_stats}, finding that they do not significantly outperform a random uncertainty score generator. We conjecture that the limited performance of all investigated UQ extensions arises from the somewhat imprecise description of potential reasoning branches inherent in the turn-level reasoning process.

Inspired by this recognition, we propose the Graph-based Reasoning UncErtainty in Trajectories (GRUET) method centering on the precise quantification of turn-level reasoning uncertainty and simple trajectory-level aggregations. Specifically, GRUET explicitly models potential branches of reasoning chains with a Directed Acyclic Graph (DAG), where each node represents a reasoning step or a textual action, and each edge points to the next reasoning step, thus precisely describing potential branches in the turn-level reasoning process. Since the collection of all potential branches, i.e., the reasoning space~\citep{chen2025towards}, covers all possible reasoning chains, we can exploit the graph complexity to approximate reasoning space complexity. We construct the graph complexity by integrating token-level distributional statistics and topological information. Next, GRUET employs simple strategies to aggregate the estimated turn-level reasoning uncertainty across multiple turns into a trajectory-level uncertainty measure, which quantifies the overall credibility of the trajectory. Figure~\ref{fig:overview_ureact} illustrates the workflow of GRUET. Empirical results across two LLM families, nine parameter scales, and five benchmarks validate the effectiveness of GRUET in terms of selective generation performance.

The rest of this paper is organized as follows. Section~\ref{sec:related_work} reviews related work. Section~\ref{sec:uq} formally introduces the GRUET method. Section~\ref{sec:experiments} conducts extensive experiments to validate the effectiveness of our proposed GRUET. Section~\ref{sec:conclusions} concludes this work.

\section{Related Work}  \label{sec:related_work}

\paragraph{UQ of LLMs} UQ methods of LLMs aim to characterize and quantify the uncertainty in a single-turn LLM generation, and can be roughly divided into three categories involving reflexive-based, information-based, and diversity-based methods~\cite{vashurin2025benchmarking}. Reflexive-based UQ methods~\cite{kadavath2022ptrue,xiong2024can} prompt the LLM to judge its own uncertainty with sophisticated templates. Information-based and diversity-based UQ methods model LLM generations as sequences, then quantify uncertainty via information-theoretic metrics~\cite{fomicheva2020ppl,malinin2021mcse} and heuristic diversity metrics~\cite{duan2024sar,farquhar2024detecting,lin2024generating}, respectively. All three categories of LLM UQ are limited in characterizing the potential branches that emerge at each reasoning step when quantifying reasoning uncertainty.

\paragraph{UQ of LLM-driven Agents} An intuitive approach to quantifying the uncertainty of the ReAct process is to extend LLM UQ methods by performing them at each turn and aggregating the quantified uncertainty at the trajectory level~\cite{duan2025uprop,oh2026uncertainty}. A line of work focuses on improving turn-level UQ by extending reflexive-based methods via more sophisticated prompt templates~\cite{zhang2026confidence,zhang2026agenticUQ}. The prompt-based generation process exhibits intrinsic stochasticity induced by token-level distribution and stochastic sampling, which recursively amplifies the uncertainty in LLM generation~\cite{ou2026origins}. Thus, such reflexive-based extensions are fundamentally limited in their effectiveness. Another line of work focuses on constructing trajectory-level aggregation strategies, using weighted averaging rather than simple averaging. These weights are typically set via supervised learning, using target labels such as binary labels indicating trajectory correctness~\cite{zhang2026agenticConf}, and human-annotated ternary labels describing the degree of semantic shift of the current turn from the initial turn and that between the current observation and reasoning~\cite{zhao2025uncertainty}. Such supervised learning approaches are costly and lack plug-and-play applicability. Therefore, developing an effective and general UQ method for LLM-driven agents remains necessary and challenging.

\section{Uncertainty Quantification}  \label{sec:uq}
In this section, we propose the Graph-based Reasoning UncErtainty in Trajectories (GRUET) method to quantify the uncertainty of the ReAct process. Before that, we present some notations. Let $[N] = \{1, 2, ..., N\}$ be an integer set for $N \in \mathbb{N}^+$, and $|\cdot|$ denotes the number of elements in a collection, e.g., $|[N]|=N$. The symbol $\lfloor z \rfloor$ indicates the largest integer not exceeding $z\in \mathbb{R}$, e.g., $\lfloor 1.1 \rfloor = 1$. We denote by $x_{(i)} \in X$ the $i$-th largest element in a finite non-empty set $X$ for $i \in [|X|]$, e.g., $x_{(2)}=3$ for $X=[4]$.

We start by formalizing the procedures of ReAct, as illustrated in the left panel of Figure~\ref{fig:overview_ureact}. ReAct works with a given problem $x$ in the form of text and an initial history $\boldsymbol{h}_{0}$. At the $t$-th turn for $t \in \mathbb{N}^*$, the LLM takes the history $\boldsymbol{h}_{t-1}$ as input to generate a reasoning chain $r_t = (s_{t,1}, s_{t,2}, \dots, s_{t,n_t}, a_t)$, where $s_{t,i}$ denotes the $i$-th reasoning step and $a_t$ marks the textual action for $i \in \mathbb{N}^*$ and for $n_t \in \mathbb{N}^*$. Subsequently, the agent deterministically parses the textual action $a_t$ from the reasoning chain $r_t$ to obtain an executable action $\hat{a}_t$, executes it in the environment, and receives an observation $o_t$ in the form of text. Next, the text pair $(r_t, o_t)$ is appended to the history to form the updated history $\boldsymbol{h}_{t}=(x, r_1, o_1, \dots, r_t, o_t)$. This loop terminates when a specific action is executed, or a pre-specified maximum number of turns is reached, ending at the $T$-th turn to form the trajectory $\boldsymbol{h}_T$ for $T \in \mathbb{N}^*$. The trajectory $\boldsymbol{h}_T$ is evaluated by a reward function $\mathcal{R}$, yielding a reward $\rho = \mathcal{R}(\boldsymbol{h}_T) \in \mathbb{R}$; this reward is typically binary in code-generation benchmarks such as the SWE-bench series~\cite{jimenez2024swebench}, where $\rho=0$ and $\rho=1$ indicate incorrect and correct trajectories, respectively.

GRUET aims to provide an uncertainty score $u \in \mathbb{R}$ for the trajectory $\boldsymbol{h}_T$, with the key idea of precisely quantifying the turn-level reasoning uncertainty via graphs and selectively aggregating these turn-level uncertainty scores to provide a trajectory-level uncertainty. This idea is illustrated in Figure~\ref{fig:overview_ureact} and detailed in the following paragraphs.

\subsection{Turn-level Reasoning UQ}  \label{subsec:turn-level_UQ}
This subsection describes the turn-level reasoning UQ procedure of the proposed GRUET, which aims to quantify the uncertainty $u_t$ of the reasoning chain $r_t$. We recognize that uncertainty in LLM reasoning often manifests as divergent branches emerging at each reasoning step, even when the LLM is fed the same input. Inspired by this recognition, we model the reasoning process as a DAG, where each node denotes a reasoning step or a textual action and each edge points to the next reasoning step or action, and implement this modeling by performing multiple sampling given the same history $\boldsymbol{h}_{t-1}$ and merging equivalent nodes to reveal the intrinsic topology of the reasoning space spanned by all possible reasoning branches. Intuitively, a flawed reasoning process would be highly uncertain and exhibit many potential branches at each step, thus inducing a complex topology of the reasoning space. Hence, we can quantify the reasoning uncertainty by the complexity of the reasoning space, which can be approximated by the graph complexity. This graph-based reasoning UQ achieves a precise characterization of the reasoning process, thereby enabling a precise quantification of reasoning uncertainty. The turn-level reasoning UQ procedure comprises stages of graph construction, step-level uncertainty quantification, and graph complexity estimation, which are detailed as follows.

\paragraph{Graph Construction} This stage constructs a DAG to describe potential reasoning branches in the reasoning space. The key idea is to perform multiple sampling, model each reasoning step or textual action as a node, connect nodes with edges pointing to the next step, and merge semantically equivalent nodes to reveal the intrinsic topology of the reasoning space. Specifically, we first draw $K-1$ samples to obtain candidate reasoning chains 
\[
    r_t^k=(s_{t,1}^k, \dots, s_{t,n_{t,k}}^k, a_t^k), \ \text{for} \ K \geq 2 \ \text{and} \ k \in [K-1] \ .
\]
$r_t$ is also denoted as $r_t^K$ for the simplicity of indexing. Next, we initialize the DAG by treating the trajectory history $\boldsymbol{h}_{t-1}$, each reasoning step $s_{t,j}^k$, and each textual action $a_t^k$ as nodes, then adding directed edges from $\boldsymbol{h}_{t-1}$ to $s_{t,1}^k$, from $s_{t,j}^k$ to $s_{t,j+1}^k$ for $j \in [ n_{t,k} - 1 ]$, and from $s_{t,n_{t,k}}^k$ to $a_t^k$. The ``Graph Initialization'' column of Figure~\ref{fig:overview_ureact} illustrates the above procedure. Finally, we employ two distinct strategies to judge the semantic equivalence of reasoning nodes and action nodes, respectively. Since reasoning nodes usually contain plain natural language, we merge two reasoning nodes if an NLI model indicates that each node entails the other, a criterion known as bidirectional entailment~\cite{farquhar2024detecting}. By contrast, action nodes often contain task-specific content like code snippets. We therefore merge two action nodes if the cosine similarity between their embeddings, obtained via task-specific embedding models, exceeds a pre-specified threshold $\tau \in \mathbb{R}^+$. Algorithm~\ref{alg:AOV} in Appendix~\ref{app:subsec:GUT-Q} lists the above graph construction procedures.

\begin{algorithm}[t!]
\caption{Graph Complexity Estimation}
\label{alg:DAG_UP}
\textbf{Input:} DAG $G=(V,E)$, node uncertainty $\{U(v)\}_{v \in V}$, weight $\omega \in \mathbb{R}^+$, and activation function $\phi$. \\
\textbf{Output:} Graph complexity $\text{GC}(G)$.\\
\textbf{Procedures:}
\begin{algorithmic}[1]
\STATE Construct graph $G'=(V', E')$ by connecting all leaf nodes of $G$ to a new auxiliary node $s_a$
\STATE Construct an FNN corresponding to $G'$ with $\phi$
\STATE Set weights $w_{uv} \gets \omega$ for all $(u, v) \in E'$
\STATE Set biases $b_v \gets -1 / U(v)$ for all $v \in V$, and $b_{s_a} \gets 0$
\STATE $\textrm{GC}(G) \gets$ output of $s_a$ from forward propagation
\end{algorithmic}
\end{algorithm}

\paragraph{Step-level Uncertainty Quantification} This stage quantifies step-level uncertainty for capturing inherent stochasticity arising from token-level distributions during the process of LLM generation~\cite{ou2026origins}. The key idea is to calculate token-level uncertainty~\citep{shorinwa2025survey} and aggregate them to get step-level uncertainty. Let $s=(c_1,\dots,c_{|s|})$ be any reasoning step comprising $|s|\in \mathbb{N}^*$ tokens, each token $c_j$ has a token-level distribution $\pi( \cdot | c_{<j})$ over the vocabulary $\mathcal{V}$, where $c_{<j}$ denotes the context $(c_1,\dots,c_{j-1})$ for $j \in \mathbb{N}^*$. Drawing inspiration from~\citet{fu2025deep}, we calculate the step-level uncertainty and set hyperparameters according to their guidelines as follows.
\begin{itemize}
    \item First, uncertainty $U(c_j) = \sum\nolimits_{c\in \mathcal{V}}^{}{\log \pi_{\theta}(c | h_{<j})} / | \mathcal{V} |$ is computed for the token $c_j$, where $j \in [|r|]$.
    \item Second, we calculate $U_i =\sum_{j=i}^{i+w-1}{U( c_j )} / w$ to get a set $\{U_i\}_{i \in [|s|-w+1]}$ for $w \in [|s|]$ to capture the local uncertainty of consecutive tokens. We set $w = \min (4, |s|)$.
    \item Third, step-level uncertainty $U(s)$ is obtained by averaging the top-$d\%$ values among $\{U_i\}_{i \in [|s|-w+1]}$, i.e., $\{U_{(i)}\}_{i \in [\beta ]}$ for $\beta = \max (1, \lfloor (|s| - w + 1 ) d\% \rfloor )$, where $d$ is set to $10$.
\end{itemize}
Since each reasoning step $s$ is modeled as a node $v$, we can term step-level uncertainty $U(s)$ node uncertainty, denoted as $U(v)$.

\paragraph{Graph Complexity Estimation} This stage constructs graph complexity by integrating both the statistics of token-level distributions and topological information encoded in the constructed DAG. Since a DAG can be viewed as a Feedforward Neural Network (FNN)~\citep{scarselli2008graph} and uncertainty accumulates along reasoning steps~\citep{gan2025rethinking,zhang2024how}, one can propagate node uncertainty from the first reasoning steps to an auxiliary node $s_a$ where all textual actions are directed. Algorithm~\ref{alg:DAG_UP} summarizes the above idea and lists the calculation procedure. Implementation details are provided in Appendix~\ref{app:subsec:GUT-Q}.

\begin{figure*}[t!]
    \centering
    \includegraphics[width=\linewidth]{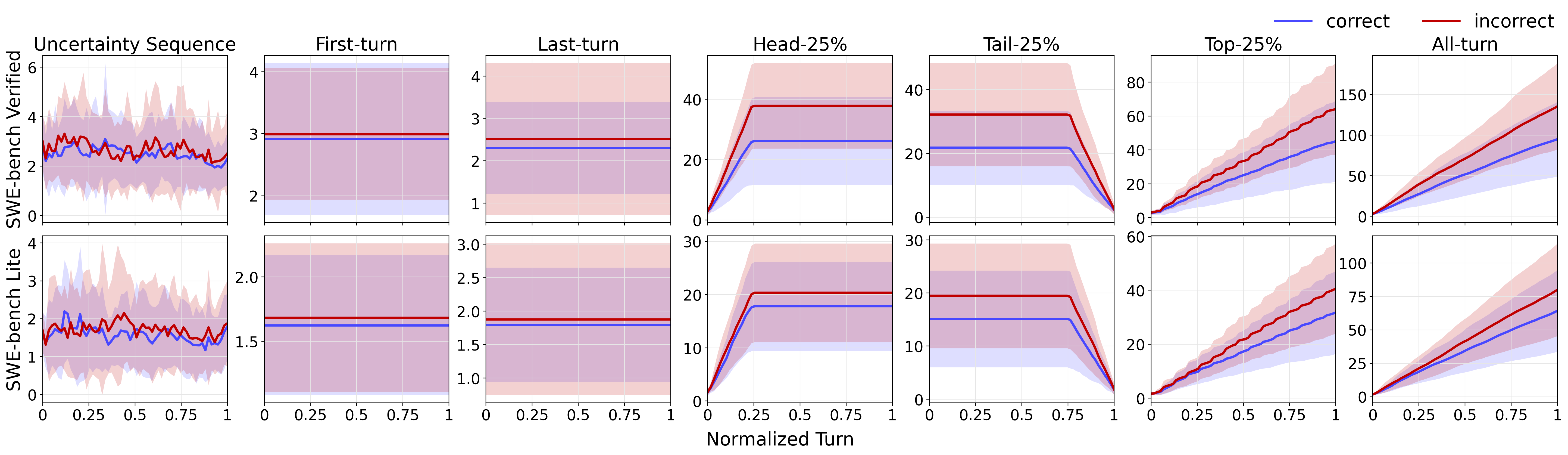}
    \caption{Visualizations of aggregation strategies on SWE-bench Verified and Lite benchmarks with the Qwen3.5-35B-A3B.}
    \label{fig:comparison_aggregation_qwen35-35b}
\end{figure*}

\subsection{Trajectory-level Uncertainty Aggregation}  \label{subsec:trajectory-level_aggregation}
This subsection proposes strategies to provide a trajectory-level uncertainty by selectively aggregating turn-level uncertainty scores. The first column of Figure~\ref{fig:comparison_aggregation_qwen35-35b} visualizes the uncertainty sequence $\boldsymbol{u}_L$ of the Qwen3.5-35B-A3B on the SWE-bench Verified and Lite benchmarks~\cite{jimenez2024swebench}, where the line and region separately denote the mean and standard deviation, and blue and red represent correct and incorrect trajectories, respectively. We observe that the red line is usually higher than the blue one, especially in the intervals $[0, 0.1]$ and $[0.75, 1]$. This observation suggests that position and value may be two key factors in trajectory-level aggregation for discriminating between correct and incorrect trajectories; position refers to the head or tail of the uncertainty sequence $\boldsymbol{u}_L$, whereas value denotes the top elements within the uncertainty set $\{u_1,\dots,u_L\}$. Inspired by this insight, we propose several aggregation strategies, as summarized in Table~\ref{tab:aggregation_strategy}. The first-turn, last-turn,  head-$d\%$, and tail-$d\%$ strategies are position-based, motivated by the belief that uncertainty scores at specific trajectory positions can effectively discriminate between correct and incorrect trajectories. By contrast, the top-$d\%$ strategy is value-based, motivated by the belief that the most uncertain turns are the most informative for such discrimination. The all-turn strategy simply averages the uncertainty scores across all turns.

\begin{table}[ht]
\footnotesize
\centering
\begin{tabular}{lll}
\toprule
Strategy   & Formulation                                                                                                         & Number of Samples\\
\midrule
First-turn & $u = u_1$                                                                                                           & $K-1$          \\
Last-turn  & $u = u_T$                                                                                                           & $K-1$          \\
Head-$d\%$ & \begin{tabular}{@{}l@{}}$u = \sum_{i=1}^\alpha {u_i}/ \alpha$ for $\alpha = \max ( \lfloor T d\% \rfloor , 1 )$\end{tabular} & $\alpha(K-1)$  \\
Tail-$d\%$ & $u = \sum_{i=T-\alpha+1}^T {u_i}/ \alpha$                                                                           & $\alpha(K-1)$  \\
Top-$d\%$  & $u = \sum_{i=1}^\alpha u_{(i)} / \alpha$                                                                            & $T(K-1)$       \\
All-turn   & $u = \sum_{i=1}^T{u_i}/T$                                                                                           & $T(K-1)$       \\
\bottomrule
\end{tabular}
\caption{Formulation and number of samples of the proposed aggregation strategies.}
\label{tab:aggregation_strategy}
\end{table}

Figure~\ref{fig:comparison_aggregation_qwen35-35b} shows the visualizations of the proposed aggregation strategies on SWE-bench Verified and Lite benchmarks with the Qwen3.5-35B-A3B. It is obvious that the blue line and region can be separated from their red counterparts for all six proposed aggregation strategies. This observation indicates that simple statistics extracted from the uncertainty sequence $\boldsymbol{u}_L$ can be used to discriminate between correct and incorrect trajectories, which validates the effectiveness of our proposed aggregation strategies. Numerical comparisons of the discriminative performance of these aggregation strategies are provided in Section~\ref{sec:experiments} and Appendix~\ref{app:subsec:aggregation_strategies}.

\section{Experiments}  \label{sec:experiments}
In this section, we conduct experiments to validate the effectiveness of the proposed GRUET method. The experiments are performed to answer (Q1) whether and to what extent the proposed GRUET outperforms classical UQ methods in discriminating between correct and incorrect trajectories; and (Q2) how the choice of aggregation strategies influences the performance of our proposed GRUET.

\paragraph{Configurations} Experiments were conducted on NVIDIA A100 80GB GPUs ($\times$4). The evaluated models include the Qwen3.5\footnote{https://huggingface.co/collections/Qwen/qwen35} family at five sizes, namely 35B, 27B, 9B, 4B, and 2B, and the Gemma4\footnote{https://huggingface.co/collections/google/gemma-4} family at four sizes, namely 31B, 26B, 8B, and 5B. The configuration of sampling parameters is detailed in Appendix~\ref{app:subsec:UQ_hyperparameter}. For auxiliary models, we use the Deberta-large~\citep{he2021deberta} that is fine-tuned on the MNLI~\citep{williams2018broad} dataset as the NLI model, and the CodeXEmbed model~\citep{liu2025codexembed} as the embedding model. Our methods were evaluated on agentic code generation benchmarks of varying difficulty, including the more challenging SWE-bench~\citep{jimenez2024swebench} series and the simpler InterCode~\citep{yang2023intercode} series. The SWE-bench series includes SWE-bench Verified\footnote{https://www.swebench.com/verified.html} and SWE-bench Lite\footnote{https://www.swebench.com/lite.html}. We use the SWE-agent~\citep{yang2024sweagent} as the agentic execution framework. The InterCode series includes the InterCode MBPP, InterCode Spider, and InterCode NL2Bash benchmarks, which adopt the InterCode as the agentic framework to separately execute the MBPP~\citep{austin2021program}, Spider~\citep{yu2018spider}, and NL2Bash~\citep{lin2018nl2bash} benchmarks. Details regarding the evaluated benchmarks are provided in Appendix~\ref{app:subsec:benchmarks}.

We selected ten representative black-box and white-box turn-level UQ methods of various categories, including the reflexive-based methods such as P(True)~\citep{kadavath2022ptrue} and Verbalized Confidence (VC)~\citep{xiong2024can}, information-based methods like Perplexity (Ppl)~\citep{fomicheva2020ppl} and Predictive Entropy (PE)~\citep{malinin2021mcse}, and the diversity-based methods including Eccentricity (Ecc), Sum of Eigenvalues (Eig), Degree matrix (Deg) of Graph Laplacian~\citep{lin2024generating}, Semantic Density (SD)~\citep{qiu2024semantic}, Sentence SAR (SSAR)~\citep{duan2024sar}, and Semantic Entropy (SE)~\citep{farquhar2024detecting}. Moreover, we consider the UQ baseline that randomly samples uncertainty scores uniformly from the interval $[0, 1]$, denoted ``Random (Rnd)''. Following previous work~\citep{duan2025uprop,oh2026uncertainty}, each contender's turn-level uncertainty scores are averaged across turns to obtain the trajectory-level uncertainty. Details regarding the concerned contenders are provided in Appendix~\ref{app:subsec:UQ_contenders}. 

\begin{figure*}[t!]
    \centering
    \includegraphics[width=0.99\linewidth, height=\textheight, keepaspectratio]{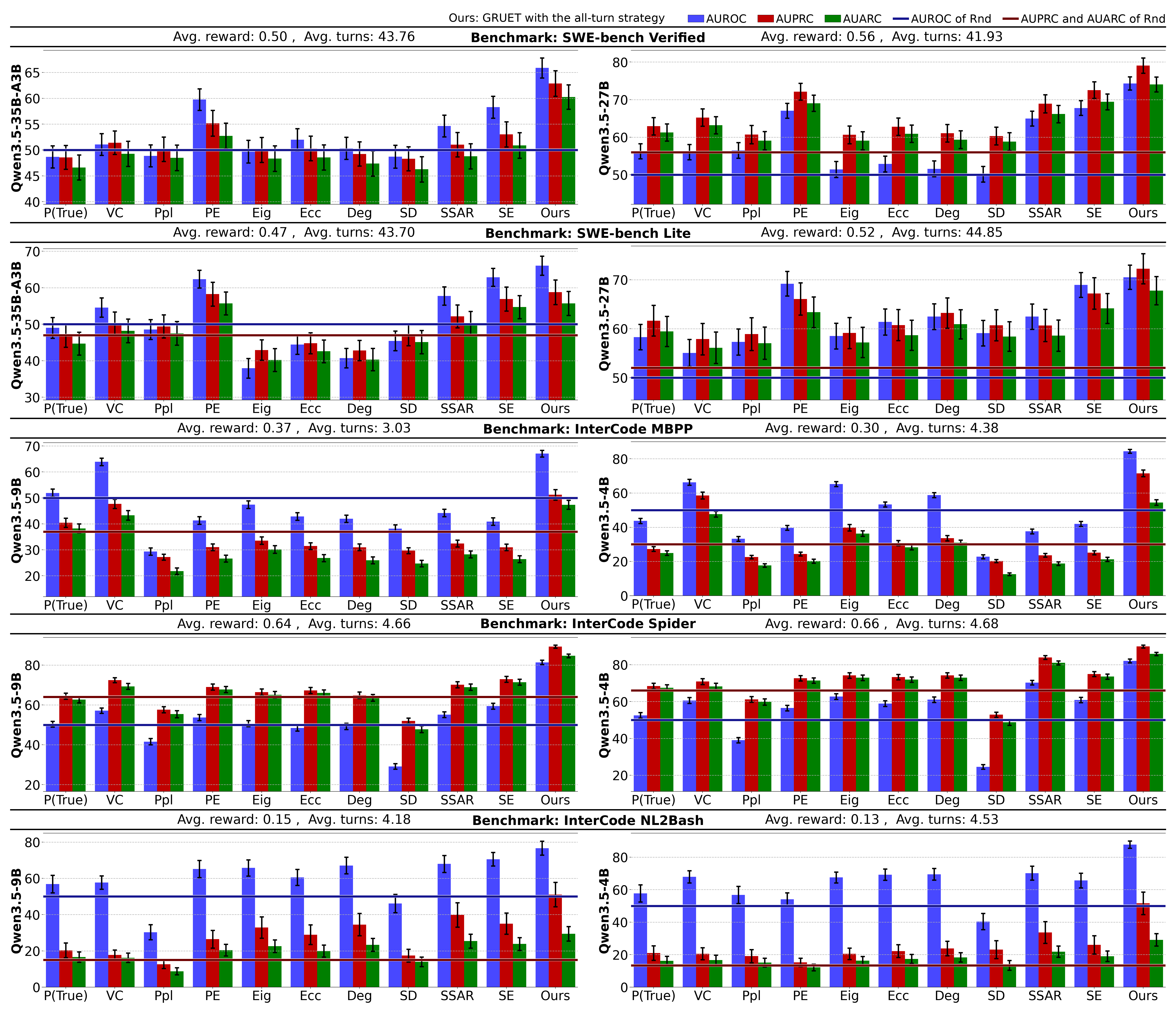}
    \caption{Comparisons of UQ performance for partial Qwen3.5 family.}
    \label{fig:uq_qwen35}
\end{figure*}

Following seminal studies~\cite{farquhar2024detecting,lin2024generating}, we evaluate a UQ method by its performance on selective generation~\citep{ren2023out}, which reflects the UQ method's ability to discriminate between correct and incorrect trajectories. Such ability is quantified using three metrics, namely the Area Under the Receiver Operating Characteristic (AUROC), the Area Under the Precision-Recall Curve (AUPRC), and the Area Under the Accuracy-Rejection Curve (AUARC)~\citep{nadeem2009accuracy}. For all three metrics, higher values indicate better discriminative performance. Following~\citet{vashurin2025benchmarking}, we bootstrap datasets 1000 times and report the mean and standard deviation (std) of the metrics. It is notable that the AUROC of Rnd is theoretically constant at $0.5$, while its AUPRC and AUARC remain constant and equal to the average (avg) reward on a given benchmark.

\paragraph{Verification on the UQ Performance} Figure~\ref{fig:uq_qwen35} shows the comparisons of UQ performance for partial Qwen3.5 family, where the blue, red, and green bars separately denote AUROC, AUPRC, and AUARC values, the blue line denotes the AUROC of Rnd, and the red line denotes both the AUPRC and AUARC of Rnd. Results of all evaluated LLMs are provided in Appendix~\ref{app:subsec:uq_performance}. We have three key observations. First, the blue bars of Ppl, P(True), and SD are typically close to the blue line, and their red and green bars are typically close to the red line. This observation indicates that the Ppl, P(True), and SD contenders usually exhibit near-random discriminative performance. Second, the bars for VC, PE, Eig, Ecc, Deg, SSAR, and SE are often close to, but sometimes significantly higher than, the corresponding lines. This observation indicates that VC, PE, Eig, Ecc, Deg, SSAR, and SE can exhibit effective discriminative ability in some cases, but such effectiveness is not general. Thus, their effectiveness is limited. Third, the bars of our proposed GRUET are consistently and significantly higher than both the corresponding lines and the bars of all contenders. This observation validates the effectiveness and generality of our proposed GRUET. Similar observations and conclusions also hold for the Gemma 4 family, as shown in Figure~\ref{fig:uq_gemma4_app} and analyzed in Appendix~\ref{app:subsec:uq_performance}.

\begin{table*}[t!]
  \centering
  \footnotesize
  \setlength{\tabcolsep}{3.75pt}
  \resizebox{\linewidth}{!}{
    \begin{tabular}{ccccccccccccccc}
      \toprule
      \multicolumn{2}{c}{\multirow{2}{*}{Metric}}
      & \multirow{2}{*}{Rnd}
      & \multicolumn{2}{c}{Reflexive-based}
      & \multicolumn{2}{c}{Information-based}
      & \multicolumn{6}{c}{Diversity-based}
      & \multicolumn{2}{c}{GRUET (Ours)} \\
      \cmidrule(lr){4-5} \cmidrule(lr){6-7} \cmidrule(lr){8-13} \cmidrule(lr){14-15}
        &           &         & P(True) & VC      & Ppl     & PE      & Eig     & Ecc     & Deg     & SD      & SSAR    & SE      & First-turn          & All-turn       \\
      \midrule
      \multirow{2}{*}{AUROC}
        & Avg. mean & $50.00$ & $50.07$ & $55.67$ & $48.20$ & $56.89$ & $55.74$ & $55.06$ & $55.04$ & $46.05$ & $54.03$ & $58.38$ & $\underline{66.11}$ & \textbf{75.49} \\
        & Avg. std. & $0.00$  & $2.53$  & $2.25$  & $2.36$  & $2.35$  & $2.30$  & $2.27$  & $2.28$  & $2.37$  & $2.39$  & $2.30$  & $2.31$              & $1.93$         \\[0.2em]

      \multirow{2}{*}{AUPRC}
        & Avg. mean & $36.57$ & $39.34$ & $41.92$ & $37.86$ & $42.75$ & $43.07$ & $42.35$ & $42.29$ & $37.49$ & $41.97$ & $44.68$ & $\underline{54.05}$ & \textbf{61.10} \\
        & Avg. std. & $0.00$  & $2.37$  & $2.23$  & $2.30$  & $2.55$  & $2.68$  & $2.63$  & $2.67$  & $2.51$  & $2.63$  & $2.74$  & $3.18$              & $3.20$         \\[0.2em]

      \multirow{2}{*}{AUARC}
        & Avg. mean & $36.57$ & $36.75$ & $39.05$ & $34.56$ & $38.98$ & $38.77$ & $37.96$ & $37.82$ & $33.02$ & $37.67$ & $40.13$ & $\underline{47.14}$ & \textbf{51.91} \\
        & Avg. std. & $0.00$  & $2.15$  & $2.23$  & $2.11$  & $2.20$  & $2.29$  & $2.24$  & $2.25$  & $2.18$  & $2.18$  & $2.25$  & $2.47$              & $2.44$         \\
      \bottomrule
    \end{tabular}
  }
  \caption{Comparisons of the overall discriminative performance of GRUET and its contenders, averaged across all evaluated LLMs and benchmarks, where bold and underlined values denote the best and second-best results, respectively.}
  \label{tab:uq_performance_stats}
\end{table*}

Table~\ref{tab:uq_performance_stats} presents the comparisons of the overall discriminative performance of contenders and GRUET, averaged across all evaluated LLMs and benchmarks, where bold and underlined values denote the best and second-best results, respectively. According to the $3\sigma$ rule in statistics~\cite{moore2017introduction}, a UQ method exhibits statistically significant non-random discriminative performance if its mean AUROC, AUPRC, and AUARC each exceed that of Rnd by more than three standard deviations. All contenders fail to meet this rule, including reflexive-based, information-based, and diversity-based UQ methods, indicating that they generally exhibit near-random discriminative performance. This finding extends the conclusions of~\citet{oh2026uncertainty} by incorporating diversity-based UQ methods. By contrast, our proposed GRUET meets this rule and outperforms the strongest contender with average improvements of 17.11, 16.42, and 11.78 in AUROC, AUPRC, and AUARC, respectively. This observation validates the effectiveness of our proposed GRUET, thereby answering Q1.

\paragraph{Impact of Aggregation Strategies} Table~\ref{tab:aggregation_strategy_impact_fine} shows the comparisons of the average discriminative performance of all proposed aggregation strategies with the Qwen3.5 and Gemma4 families. Results for each evaluated LLM are provided in Appendix~\ref{app:subsec:aggregation_strategies}. There are two key observations. First, the all-turn and Top-$25\%$ aggregation strategies generally yield the most competitive discriminative performance among all strategies, which demonstrates their generality and effectiveness. Second, the first-turn strategy usually exhibits performance comparable to that of the all-turn strategy across metrics and benchmarks. As shown in Table~\ref{tab:aggregation_strategy}, the first-turn strategy requires only $K-1$ additional samples, which is only $1/L$ of the sampling cost of the all-turn strategy. Thus, the first-turn strategy achieves a favorable trade-off between performance and sampling cost, shedding light on a lightweight implementation of the proposed GRUET. These two observations and the corresponding conclusions answer Q2.

\begin{table*}[t!]
  \centering
  \setlength{\tabcolsep}{0.46pt}
  \resizebox{\linewidth}{!}{
    \begin{tabular}{cccccccccccccccc}
      \toprule
      \multirow{2}{*}{Strategy} & \multicolumn{3}{c}{SWE-bench Verified} & \multicolumn{3}{c}{SWE-bench Lite} & \multicolumn{3}{c}{InterCode MBPP} & \multicolumn{3}{c}{InterCode Spider} & \multicolumn{3}{c}{InterCode NL2Bash} \\
      \cmidrule(lr){2-4} \cmidrule(lr){5-7} \cmidrule(lr){8-10} \cmidrule(lr){11-13} \cmidrule(lr){14-16}
                 & AUROC               & AUPRC               & AUARC               & AUROC               & AUPRC               & AUARC               & AUROC               & AUPRC               & AUARC               & AUROC               & AUPRC               & AUARC               & AUROC               & AUPRC               & AUARC               \\
      \midrule
      \multicolumn{16}{c}{\textbf{LLM Family: Qwen3.5}} \\
      \midrule
      First-turn & $55.36$             & $60.29$             & $58.26$             & $49.95$             & $53.92$             & $51.88$             & $66.14$             & $53.40$             & $46.60$             & $69.09$             & $77.98$             & $76.03$             & $78.41$             & $43.72$             & $26.63$             \\
      Last-turn  & $59.78$             & $59.74$             & $57.46$             & $54.85$             & $54.74$             & $52.86$             & $67.89$             & $44.37$             & $38.63$             & $59.33$             & $68.88$             & $67.34$             & $74.03$             & $28.90$             & $21.39$             \\
      Head-25\%  & $67.82$             & $69.22$             & $65.64$             & $59.95$             & $55.51$             & $53.58$             & $62.77$             & $46.89$             & $42.64$             & $68.22$             & $79.05$             & $76.93$             & $78.10$             & $47.13$             & $27.16$             \\
      Tail-25\%  & $65.17$             & $66.89$             & $63.72$             & $60.22$             & $56.97$             & $55.07$             & $70.38$             & $47.95$             & $40.96$             & $66.63$             & $75.55$             & $73.61$             & $71.86$             & $26.75$             & $20.29$             \\
      Top-25\%   & $\underline{68.66}$ & $\textbf{71.97}$    & $\textbf{67.46}$    & $\textbf{69.54}$    & $\textbf{66.20}$    & $\textbf{62.28}$    & $\underline{70.49}$ & $\underline{57.52}$ & $\underline{49.15}$ & $\textbf{83.22}$    & $\textbf{90.02}$    & $\textbf{85.59}$    & $\underline{81.07}$ & $\underline{47.97}$ & $\underline{28.24}$ \\
      All-turn   & $\textbf{70.10}$    & $\underline{70.96}$ & $\underline{67.15}$ & $\underline{68.28}$ & $\underline{65.51}$ & $\underline{61.75}$ & $\textbf{75.72}$    & $\textbf{61.35}$    & $\textbf{50.92}$    & $\underline{81.73}$ & $\underline{89.63}$ & $\underline{85.27}$ & $\textbf{82.21}$    & $\textbf{51.34}$    & $\textbf{29.29}$    \\
      \midrule
      \multicolumn{16}{c}{\textbf{LLM Family: Gemma 4}} \\
      \midrule
      First-turn & $62.98$             & $44.04$             & $38.73$             & $58.36$             & $33.80$             & $30.81$             & $\underline{71.76}$ & $\underline{54.93}$ & $\underline{48.41}$ & $71.80$             & $71.98$             & $68.05$             & $72.40$             & $48.11$             & $33.10$             \\
      Last-turn  & $63.03$             & $42.23$             & $36.41$             & $57.74$             & $39.33$             & $30.71$             & $63.50$             & $48.53$             & $44.07$             & $61.58$             & $56.88$             & $53.79$             & $68.46$             & $33.66$             & $26.26$             \\
      Head-25\%  & $65.17$             & $43.84$             & $37.55$             & $68.48$             & $38.71$             & $33.07$             & $70.91$             & $54.63$             & $48.16$             & $67.53$             & $67.78$             & $64.67$             & $72.77$             & $46.91$             & $32.80$             \\
      Tail-25\%  & $67.94$             & $39.99$             & $35.93$             & $70.45$             & $42.63$             & $36.62$             & $63.26$             & $48.45$             & $44.00$             & $62.08$             & $61.39$             & $58.09$             & $69.89$             & $34.53$             & $26.83$             \\
      Top-25\%   & $\underline{74.09}$ & $\underline{49.73}$ & $\underline{42.08}$ & $\underline{74.66}$ & $\underline{47.64}$ & $\underline{38.16}$ & $71.66$             & $54.05$             & $48.06$             & $\textbf{78.93}$    & $\textbf{79.21}$    & $\textbf{72.96}$    & $\underline{78.19}$ & $\textbf{50.94}$    & $\underline{34.83}$ \\
      All-turn   & $\textbf{75.60}$    & $\textbf{50.25}$    & $\textbf{42.20}$    & $\textbf{75.42}$    & $\textbf{48.94}$    & $\textbf{38.69}$    & $\textbf{73.58}$    & $\textbf{56.13}$    & $\textbf{49.44}$    & $\underline{77.36}$ & $\underline{76.92}$ & $\underline{71.56}$ & $\textbf{79.54}$    & $\underline{50.21}$ & $\textbf{34.88}$    \\
      \bottomrule
    \end{tabular}
  }
  \caption{Comparisons of the average discriminative performance of all proposed aggregation strategies with the Qwen3.5 and Gemma 4 families, where bold and underlined values denote the best and second-best results, respectively.}
  \label{tab:aggregation_strategy_impact_fine}
\end{table*}

\paragraph{Sensitivity Analyses} The proposed GRUET involves four key hyperparameters, namely the number of samples per turn $K$, the embedding similarity threshold $\tau$ for merging action nodes, the sampling temperature $T$, and the percentage $d\%$ for trajectory-level uncertainty aggregation. Next, we analyze the impact of $K$, $\tau$, and $T$ on UQ performance and recommend appropriate and unified configurations for the Qwen3.5 family, deferring the analysis of the factor $d\%$ and those of the Gemma 4 family to Appendix~\ref{app:subsec:sensitivity}. 

\begin{figure*}[t!]
    \centering
    \includegraphics[width=\linewidth]{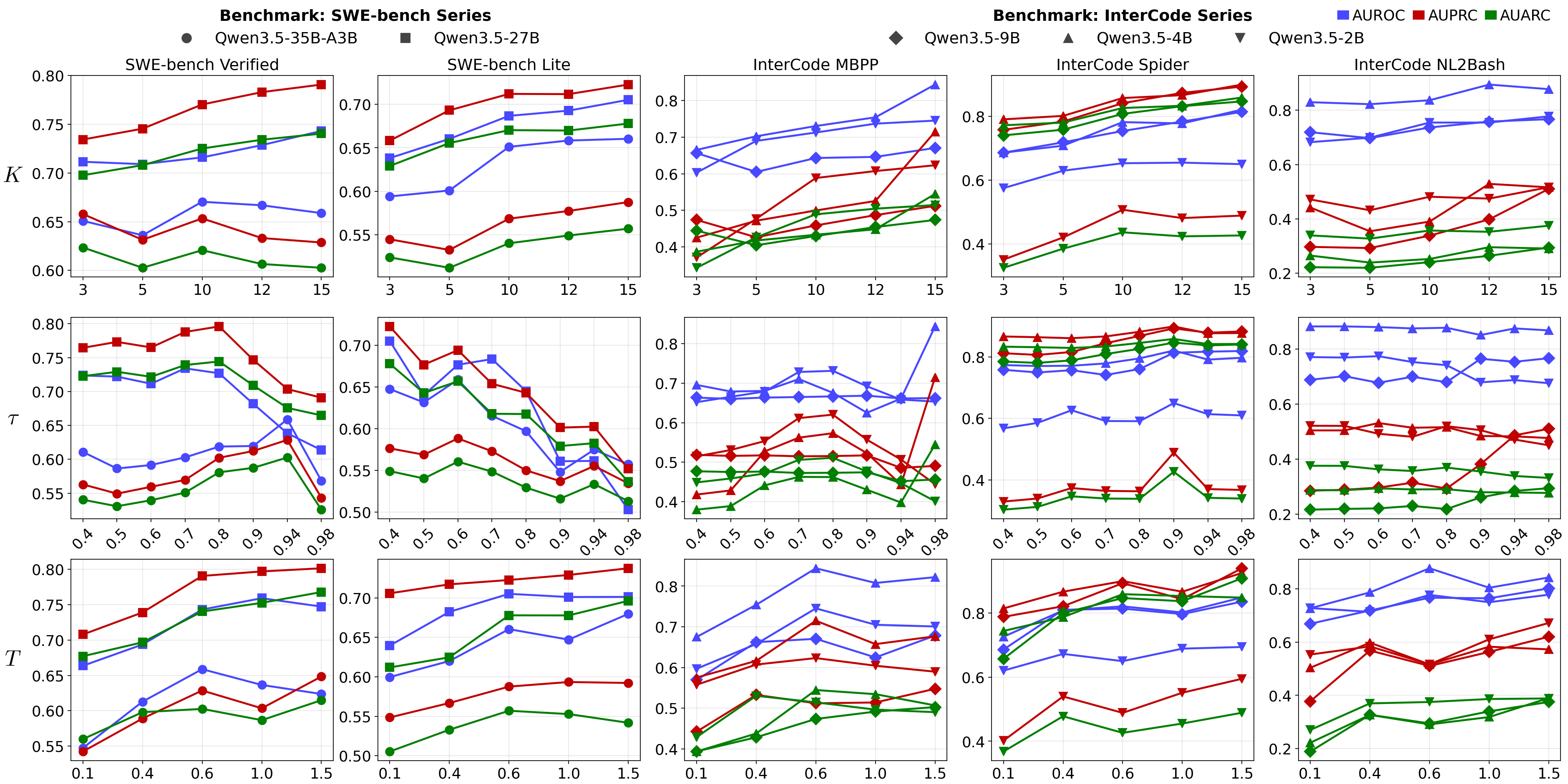}
    \caption{Sensitivity analyses of GRUET with the all-turn strategy for the Qwen3.5 family.}
    \label{fig:sensitivity_qwen}
\end{figure*}

The first row of Figure~\ref{fig:sensitivity_qwen} shows the impact of the number of samples $K$ on UQ performance. All metrics generally increase rapidly as $K$ increases from 3 to 10 and tend to converge when $K$ further increases from 10 to 15. To balance sampling cost and UQ performance, we recommend setting $K=10$. The second row shows the impact of the threshold of embedding similarity $\tau$. It is observed that all metrics are relatively sensitive to changes in $\tau$ on the SWE-bench series and InterCode MBPP, while relatively insensitive on the other two benchmarks. This observation indicates that the configuration of $\tau$ is critical to UQ performance and depends on the specific tasks. Given the overall UQ performance across all benchmarks and LLMs, we recommend setting $\tau=0.7$. The third row shows the impact of the sampling temperature $T$. All metrics are relatively low at $T=0.1$, indicating that the proposed GRUET is less effective at low temperatures. This may shed light on the fact that UQ methods requiring multiple sampling are less effective at low temperatures, which coincides with the findings in~\citet{lin2024generating}. Based on the overall UQ performance across all benchmarks and LLMs, we recommend setting $T=0.6$.

\section{Conclusions}  \label{sec:conclusions}
In this paper, we conjectured that the uncertainty of the ReAct process often arises from the cumulative turn-level reasoning uncertainty induced by LLMs. Built upon this recognition, we proposed the GRUET method that precisely quantifies the turn-level reasoning uncertainty via a DAG and employs simple aggregation strategies for quantifying the overall trajectory credibility. Empirical results across nine LLMs and five benchmarks demonstrated that GRUET can yield strong performance in selective generation tasks, while extensions of reflexive-based, information-based, and diversity-based methods typically suffers from limited performance that is comparable to that of a random uncertainty score generator.

\section*{Acknowledgments}
The research was supported by the Natural Science Foundation of China (62406138, 62632005).

\newpage
\appendix

\section*{Appendix}
This appendix provides the supplementary materials for our work ``GRUET: Quantifying Uncertainty of Agentic Reasoning-and-Acting Processes'', constructed according to the corresponding sections therein.

\section{Additional Implementation Details of GRUET}  \label{app:UQ_additional_implement}
This appendix provides additional implementation details of the proposed GRUET.

\subsection{Details on Benchmarks}  \label{app:subsec:benchmarks}
In this work, we conducted experiments on agentic code generation benchmarks of varying difficulty, including the more challenging SWE-bench~\citep{jimenez2024swebench} series and the simpler InterCode~\citep{yang2023intercode} series, with the former executed using SWE-agent~\cite{yang2024sweagent} and the latter executed using InterCode itself.

\paragraph{SWE-bench} SWE-bench is an evaluation benchmark consisting of 2,294 software engineering problems drawn from real GitHub issues and corresponding pull requests across 12 popular Python repositories, with the Verified\footnote{https://www.swebench.com/verified.html} and Lite\footnote{https://www.swebench.com/lite.html} subsets being the most widely used. SWE-bench Verified is a human-filtered subset of 500 instances, in which human annotators reviewed each instance to ensure that the problem descriptions are clear, the test patches are correct, and the tasks are solvable given the available information. SWE-bench Lite is a 300-instance subset designed for lower-cost and faster evaluation. It preserves SWE-bench's overall distribution and difficulty while focusing on more self-contained functional bug fixes.

\paragraph{InterCode} InterCode is a seminal framework of interactive agentic coding, with code as actions and execution feedback as observations. It includes the InterCode MBPP, InterCode Spider, and InterCode NL2Bash benchmarks, which instantiate interactive agentic execution settings based on the static MBPP~\citep{austin2021program}, Spider~\citep{yu2018spider}, and NL2Bash~\citep{lin2018nl2bash} benchmarks.

Table~\ref{tab:benchmark_languages} lists an overview of the evaluated benchmarks. The reward of each instance from SWE-bench and InterCode is binary, with 1 indicating solved and 0 indicating unsolved, as determined by their built-in evaluation procedures.

\begin{table}[ht]
  \centering
  \resizebox{\linewidth}{!}{
    \begin{tabular}{cccccc}
      \toprule
      \multirow{2}{*}{Benchmarks}
      & \multicolumn{2}{c}{SWE-bench Series}
      & \multicolumn{3}{c}{InterCode Series} \\
      \cmidrule(lr){2-3} \cmidrule(lr){4-6}
                & SWE-bench Verified & SWE-bench Lite   & InterCode MBPP & InterCode Spider & InterCode NL2Bash \\
      \midrule
      Language  & Python             & Python           & Python         & SQL              & Bash              \\
      Level     & Repository-level   & Repository-level & Function-level & Query-level      & Command-level     \\
      Instances & 500                & 300              & 974            & 1034             & 200               \\
      Reward    & Binary             & Binary           & Binary         & Binary           & Binary            \\
      \bottomrule
    \end{tabular}
  }
  \caption{Overview of the evaluated benchmarks.}
  \label{tab:benchmark_languages}
\end{table}

The difficulty level is characterized by the average turns and accuracy, which is detailed in Figures~\ref{fig:uq_qwen35_app} and~\ref{fig:uq_gemma4_app}. We note that small-scale LLMs, namely Qwen3.5 9B, 4B, and 2B, as well as Gemma 4 E2B and E4B, failed to solve any instances on the SWE-bench series and are therefore omitted. Conversely, large-scale LLMs, namely Qwen3.5 27B and 35B-A3B, alongside Gemma 4 26B-A4B and 31B, perform particularly well on the InterCode series, typically achieving an accuracy above 0.9. This high performance leads to severe class imbalance, making them unsuitable for evaluating UQ performance in selective generation, leading to their omission. We note that the presented accuracy may differ from that of the official technical reports from Qwen3.5\footnote{https://huggingface.co/Qwen/Qwen3.5-27B} and Gemma 4\footnote{https://huggingface.co/google/gemma-4-31B-it}, since we employ different agentic execution frameworks, i.e., SWE-agent and InterCode. However, this work primarily focuses on quantifying the uncertainty in ReAct trajectories, instead of improving the accuracy. Therefore, the accuracy differences between ours and that of the official technical reports are considerably acceptable. We also note that the evaluated SWE-bench benchmarks~\cite{jimenez2024swebench} typically require extremely long trajectories around forty ReAct turns, representing a much harder UQ task compared to previous works that primarily focus on QA tasks involving only about seven turns~\cite{duan2025uprop,zhao2025uncertainty}.

\subsection{Implementation of Turn-level Reasoning UQ}  \label{app:subsec:GUT-Q}
This subsection provides additional details on the implementation of turn-level UQ in our proposed GRUET. Algorithm~\ref{alg:AOV} lists the procedure for turn-level UQ, where $\text{Cos}(\cdot,\cdot)$ denotes the cosine operator and $\text{Emb}(\cdot)$ denotes the embedding produced by an embedding model, $\omega$ is fixed at $0.1$ following~\citet{bucur2020epidemic}, and $\phi$ is set as the identity function that always returns the value that was used as its argument. Since node uncertainty $U(v)$ is typically negative, we apply the transformation $f(U(v))=-1/U(v)$ before graph propagation in line 4 of procedure 2 for preserving order and for yielding positive node uncertainty. Consequently, more potential branches result in higher accumulated positive node uncertainty, thereby yielding greater graph complexity.

\begin{algorithm}[ht]
\caption{Graph-based Turn-level Uncertainty Quantification}
\label{alg:AOV}
\footnotesize
\textbf{Procedure 1: DAG Construction}\\
\textbf{Input:} trajectory history $\boldsymbol{h}$, reasoning chain $r$, sampling times $K-1$, temperature $T$, and threshold $\tau$. \\
\textbf{Output:} DAG $G=(V, E)$ and node uncertainty $\{U(v)\}_{v \in V}$. \\
\textbf{Procedures:}
\begin{algorithmic}[1]
\STATE Initialize $G=(V, E)$ with root node $\boldsymbol{h}$ \COMMENT{Start graph initialization.}
\STATE Sample $K-1$ reasoning chains $\{r^k\}_{k \in [K-1]}$ at temperature $T$, and view $r$ as $r^K$
\FOR {$k \in [K]$}
    \STATE Parse reasoning steps $\{s_j^k\}_{j \in [n_k]}$ and textual action $a^k$ from $r^k$, and view $a^k$ as $s_{n_k + 1}^k$
    \STATE Calculate step-level uncertainty $U(s_j^k)$ and initialize $UC(s_j^k) \gets \{U(s_j^k)\}$ for all $j \in [n_k + 1]$  \COMMENT{See Section~\ref{sec:uq}.}
    \STATE $V \gets V \cup \{s_j^k\}_{j \in [n_k + 1]}$ \ , \ $E \gets E \cup \{(\boldsymbol{h}, s_1^k)\}$ \ , and \ $E \gets E \cup \{(s_j^k, s_{j+1}^k)\}_{j \in [n_k]}$
\ENDFOR
\STATE Initialize reasoning node set $V_t \gets \emptyset$ and action node set $V_a \gets \emptyset$ \COMMENT{Start node merging.}
\FOR {$k \in [K]$}
    \STATE Initialize node set $V_m \gets V_t$ for merging reasoning nodes
    \STATE $V_t \gets V_t \cup \{s_j^k\}_{j \in [n_k]}$ \ , \ $V_a \gets V_a \cup \{s_{n_k+1}^k\}$
    \FOR {$j \in [n_k]$}
        \FOR {each node $v \in V_m$}
            \IF {NLI predicts bi-entailment for $v$ and $s_j^k$}
                \STATE $UC(v) \gets UC(v) \cup UC(s_j^k)$ \COMMENT{Collect step-level uncertainty values.}
                \STATE Get the parent $p$ of $s_j^k$ where $(p, s_j^k) \in E$ and the child $c$ of $s_j^k$ where $(s_j^k, c) \in E$
                \STATE $E \gets E \cup \{(p, v), (v, c)\}$ \ , \ $V_t \gets V_t \setminus \{s_j^k\}$ \ , and \ $V_m \gets V_m \setminus \{v\}$ \COMMENT{Merge reasoning nodes.}
                \STATE $V_m \gets V_m \setminus \{u\}$ for all ancestor nodes $u$ of $v$ \COMMENT{Preserve DAG property after merge.}
                \STATE \textbf{break}
            \ENDIF
        \ENDFOR
    \ENDFOR
    \FOR {each node $v \in V_a$}
        \IF {$v \neq s_{n_k+1}^k$ \AND $\text{Cos}(\operatorname{Emb}(v), \operatorname{Emb}(s_{n_k+1}^k)) > \tau$}
            \STATE $UC(v) \gets UC(v) \cup UC(s_{n_k+1}^k)$ \COMMENT{Collect step-level uncertainty values.}
            \STATE Get the parent $p$ of $s_{n_k+1}^k$ where $(p, s_{n_k+1}^k) \in E$
            \STATE $E \gets E \cup \{(p, v)\}$ \ , \ $V_a \gets V_a \setminus \{s_{n_k+1}^k\}$ \COMMENT{Merge action nodes.}
            \STATE \textbf{break}
        \ENDIF
    \ENDFOR
\ENDFOR
\STATE $U(v) \gets \text{Average}(UC(v))$ for all $v \in V_t \cup V_a$ and update $G$ to retain only nodes in $V_t \cup V_a$
\end{algorithmic}

\vspace{0.1em}

\textbf{Procedure 2: Graph Complexity Estimation}\\
\textbf{Input:} DAG $G=(V,E)$, node uncertainty $\{U(v)\}_{v \in V}$, weight $\omega \in \mathbb{R}^+$, and activation function $\phi$. \\
\textbf{Output:} Graph complexity $\text{GC}(G)$.\\
\textbf{Procedures:}
\begin{algorithmic}[1]
\STATE Construct graph $G'=(V', E')$ by connecting all leaf nodes of $G$ to a new auxiliary node $s_a$.
\STATE Construct an FNN corresponding to $G'$ with $\phi$
\STATE Set weights $w_{uv} \gets \omega$ for all $(u, v) \in E'$
\STATE Set biases $b_v \gets -1 / U(v)$ for all $v \in V$, and $b_{s_a} \gets 0$
\STATE $\text{GC}(G) \gets$ output of $s_a$ from forward propagation
\end{algorithmic}
\end{algorithm}
\FloatBarrier

\subsection{Configurations of Hyperparameters}  \label{app:subsec:UQ_hyperparameter}
This subsection details the configurations of hyperparameters in this work.

\paragraph{Configuration of Sampling Parameters} Following prior work~\citep{farquhar2024detecting,qiu2024semantic}, we use different sampling setups for evaluating correctness and quantifying uncertainty.
\begin{itemize}
    \item \textbf{For Evaluating Correctness.} We follow the official Qwen3.5 recommendation\footnote{https://huggingface.co/Qwen/Qwen3.5-27B} using $T=0.6$, Top-$k=20$, and Top-$p=0.95$ to obtain the trajectory, which determines whether the problem is solved. Similarly, we follow the official Gemma 4 recommendation\footnote{https://huggingface.co/google/gemma-4-31B-it} using $T=1.0$, Top-$k=64$, and Top-$p=0.95$ to obtain the trajectory. The maximum number of newly generated tokens per turn, namely the max\_new\_tokens parameter, is set to 3072 for the Qwen3.5 family and 4096 for the Gemma 4 family.
    \item \textbf{For Evaluating Uncertainty.} By default, we use the same sampling setting as described above to obtain the additional candidate samples. In the ablation experiments, we change the sampling temperature $T$ only for these additional candidate samples, thereby leaving the evaluated correctness and overall accuracy unchanged. Moreover, the default number of samples is set to $K=15$.
\end{itemize}

\noindent \textbf{Configuration of ReAct.} Following~\citet{yang2023intercode} and~\citet{yang2024sweagent}, we set the maximum number of turns $M$ as $10$ and $60$ for the InterCode and SWE-bench series, respectively. The ReAct loop terminates if and only if this pre-specified $M$ is reached, or a specific action is executed.

\noindent \textbf{Configuration of GRUET.} The hyperparameters in GRUET include the embedding threshold $\tau$ and the trajectory-level uncertainty aggregation strategy. For our implementation, we set $\tau=0.7$ and utilized the all-turn strategy.

\paragraph{Regarding Prompt Templates} The prompt follows those of InterCode and SWE-agent, with a slight insertion to their system instruction for more accurately parsing the reasoning steps in graph construction. The insertions are detailed as follows, with which we can parse the ``Step'' markers to identify and extract each reasoning step. We follow InterCode and SWE-agent to use few-shot prompting with $N$ demonstrations, where $N=3$ for the InterCode MBPP, $N=5$ for the InterCode Spider and NL2Bash, and $N=1$ for SWE-agent.

\begin{mdframed}
Break down your reasoning process into small steps.

Each step should represent a single, minimal reasoning action, and each step must logically follow the previous one.

Use the following format for each step:

Step i: [Your reasoning process in one cohesive response]
\end{mdframed}

\section{Additional Experimental Results of GRUET} \label{app:UQ_additional_experiments}
This appendix provides additional experimental results of the proposed GRUET. 

\subsection{UQ Contenders}  \label{app:subsec:UQ_contenders}
This subsection introduces the details of the concerned contenders. Table~\ref{tab:abbr_of_ue_methods} lists the full names and corresponding abbreviations (Abbr.) of the concerned UQ contenders. We implemented these contenders via LM-Polygraph\footnote{https://github.com/IINemo/lm-polygraph}~\citep{vashurin2025benchmarking}.

\begin{table}[htbp]
  \centering
  \footnotesize
  \resizebox{\linewidth}{!}{$
  \begin{tabular}{cccccc}
          \toprule
          Type & Abbr. & Full Name & Category & Multi-sampling? & Paper \\
          \midrule
          \multirow{5}{*}{White-box}
          & Ppl  & Perplexity                           & \multirow{2}{*}{Information-based} & $\times$ & \citet{fomicheva2020ppl}      \\
          & PE   & Predictive Entropy                   &                                     & $\checkmark$     & \citet{malinin2021mcse}       \\[0.3em]
          & SE   & Semantic Entropy                     & \multirow{3}{*}{Diversity-based}    & $\checkmark$ & \citet{farquhar2024detecting} \\
          & SSAR & SentenceSAR                          &                                     & $\checkmark$     & \citet{duan2024sar}           \\
          & SD   & Semantic Density                     &                                     & $\checkmark$ & \citet{qiu2024semantic}       \\[0.3em]
          & P(True)   & P(True)                         &  Reflexive-based                    & $\times$ & \citet{kadavath2022ptrue}       \\
          \midrule
          \multirow{4}{*}{Black-box}
          & Eig  & Sum of Eigenvalues-NLI Score Entail. & \multirow{3}{*}{Diversity-based}    & $\checkmark$     & \citet{lin2024generating}     \\
          & Deg  & Degree Matrix-NLI Score Entail.      &                                     & $\checkmark$ & \citet{lin2024generating}     \\
          & Ecc  & Eccentricity-NLI Score Entail.       &                                     & $\checkmark$     & \citet{lin2024generating}     \\[0.3em]
          & VC   & Verbalized Confidence                & Reflexive-based                     & $\times$ & \citet{xiong2024can}          \\
          \bottomrule
        \end{tabular}
  $}
  \caption{Overview of the UQ contenders.}
  \label{tab:abbr_of_ue_methods}
\end{table}

\subsection{UQ Performance Evaluations}  \label{app:subsec:uq_performance}
This subsection provides additional details on the evaluation of UQ performance, including the Qwen3.5 family at scales of 2B, 4B, 9B, 27B, and 35B, and the Gemma 4 family at scales of 5B, 8B, 26B, and 31B.

\begin{figure}[ht]
    \centering
    \includegraphics[width=\linewidth]{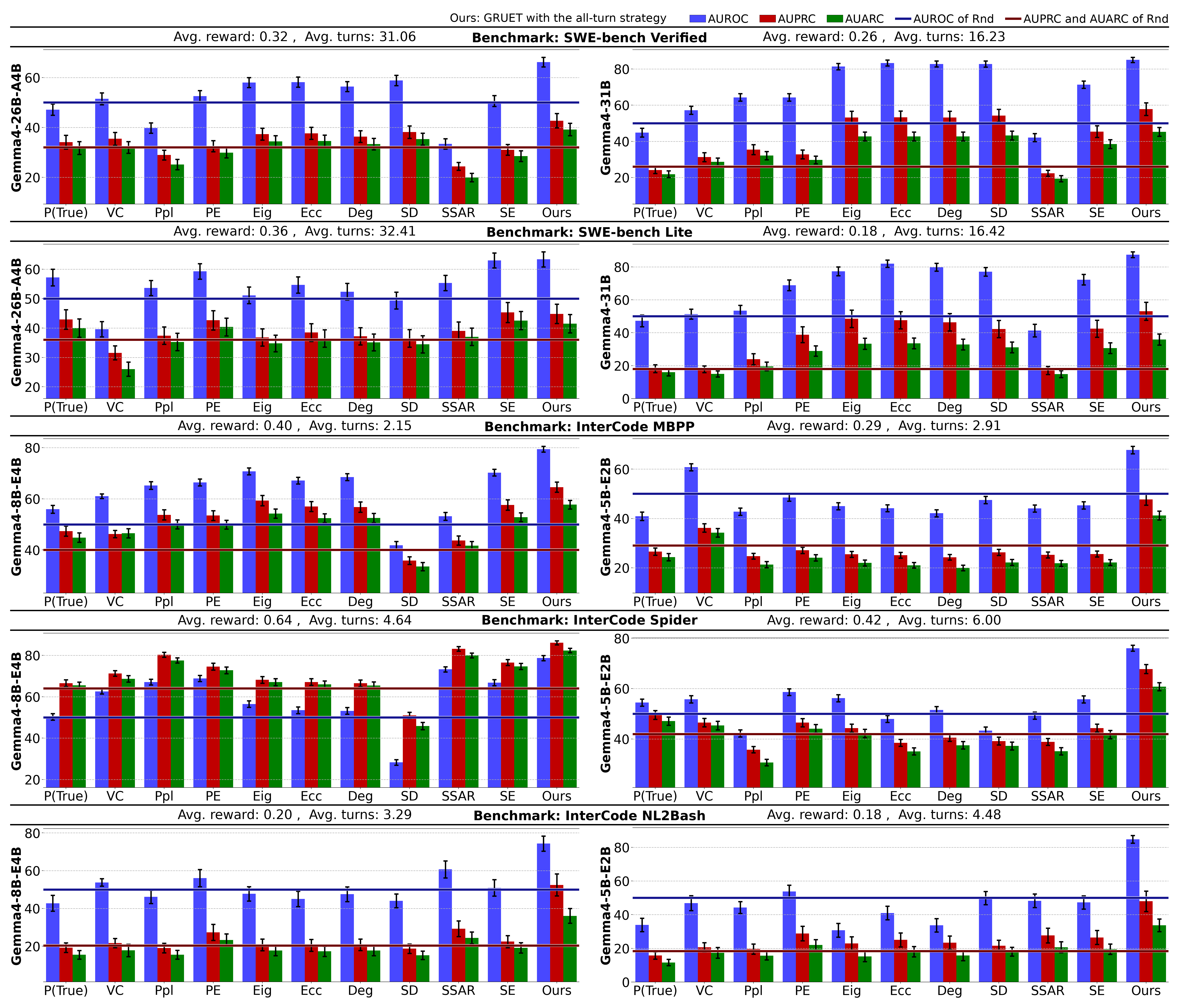}
    \caption{Comparisons of UQ performance across the Gemma 4 family.}
    \label{fig:uq_gemma4_app}
\end{figure}

\begin{figure}[ht]
    \centering
    \includegraphics[width=\linewidth, height=0.96\textheight, keepaspectratio]{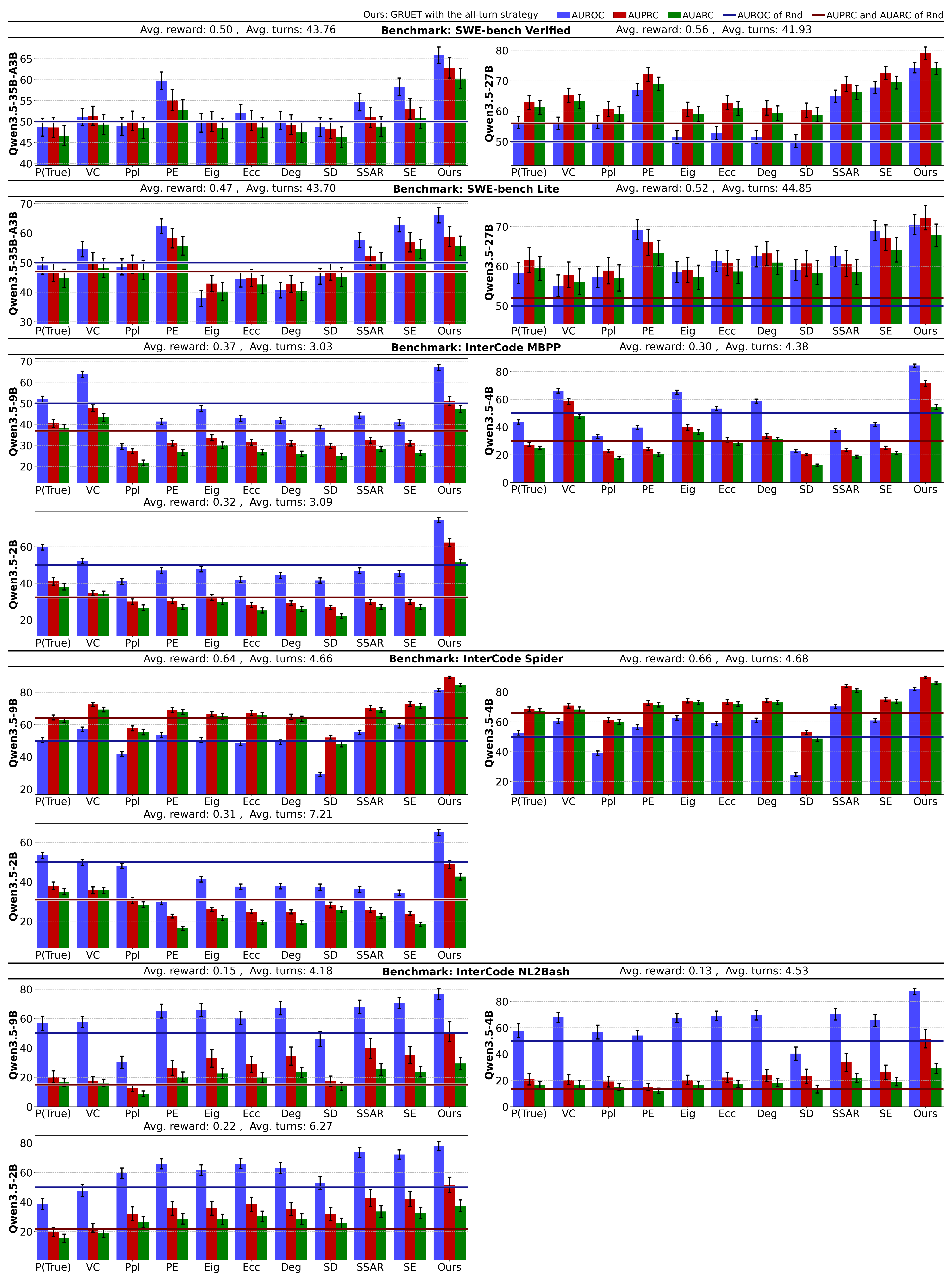}
    \caption{Comparisons of UQ performance across the Qwen3.5 family.}
    \label{fig:uq_qwen35_app}
\end{figure}

Figures~\ref{fig:uq_gemma4_app} and~\ref{fig:uq_qwen35_app} show the comparisons of UQ performance across the Gemma 4 and Qwen3.5 families, respectively. We have three key observations. First, the blue bars of Ppl, P(True), and SD are typically close to the blue line, and their red and green bars are typically close to the red line. This observation indicates that the Ppl, P(True), and SD contenders usually exhibit near-random discriminative performance. Second, the bars for VC, PE, Eig, Ecc, Deg, SSAR, and SE are often close to, but sometimes significantly higher than, the corresponding lines. This observation indicates that VC, PE, Eig, Ecc, Deg, SSAR, and SE can exhibit effective discriminative ability in some cases, but such effectiveness is not general. Thus, their effectiveness is limited. Third, the bars of our proposed GRUET are consistently and significantly higher than both the corresponding lines and the bars of all contenders. This observation validates the effectiveness and generality of our proposed GRUET. 

\clearpage
\subsection{Details on the Comparisons of Aggregation Strategies}  \label{app:subsec:aggregation_strategies}
This subsection provides additional details on the comparisons of aggregation strategies. Tables~\ref{tab:aggregation_strategy_swe_bench} and~\ref{tab:aggregation_strategy_intercode} present the comparison of the UQ performance of all proposed aggregation strategies for each LLM on the SWE-bench series and InterCode series, respectively, where bold and underlined values separately denote the best and second-best results. We have three key observations. First, the best aggregation strategy varies across UQ metrics and benchmarks, which indicates that the locations of key uncertainty scores within the uncertainty sequence $\boldsymbol{u}_L$ that correlate more strongly with correctness depend on specific evaluations and tasks. Second, the all-turn strategy usually exhibits competitive performance across UQ metrics and benchmarks, demonstrating its strong generality in discriminative performance. Third, the first-turn strategy exhibits UQ performance comparable to that of the all-turn strategy across UQ metrics and benchmarks. As shown in Table~\ref{tab:aggregation_strategy}, the first-turn strategy requires only $K-1$ samples, which is only $1/L$ of the sampling cost of the all-turn strategy. Thus, the first-turn strategy achieves a good trade-off between performance and sampling cost, shedding light on a lightweight implementation of the proposed GRUET.

\begin{table*}[ht]
  \centering
  \footnotesize
  \setlength{\tabcolsep}{2pt}
  \resizebox{\linewidth}{!}{
  \begin{tabular}{cccccccc@{\hspace{0.8em}}ccccccc}
    \toprule
    \multicolumn{8}{c}{\textbf{LLM Family: Qwen3.5}} & \multicolumn{7}{c}{\textbf{LLM Family: Gemma 4}} \\
    \cmidrule(lr){1-8} \cmidrule(lr){9-15}
    \multirow{2}{*}{Scale} & \multirow{2}{*}{Strategy} & \multicolumn{3}{c}{SWE-bench Verified} & \multicolumn{3}{c}{SWE-bench Lite} & \multirow{2}{*}{Scale} & \multicolumn{3}{c}{SWE-bench Verified} & \multicolumn{3}{c}{SWE-bench Lite} \\
    \cmidrule(lr){3-5} \cmidrule(lr){6-8} \cmidrule(lr){10-12} \cmidrule(lr){13-15}
                             &            & AUROC               & AUPRC               & AUARC               & AUROC               & AUPRC               & AUARC               &                          & AUROC               & AUPRC               & AUARC               & AUROC               & AUPRC               & AUARC               \\
    \midrule
    \multirow{6}{*}{\shortstack{35B-\\A3B}} & First-turn & $52.24$             & $58.10$             & $55.61$             & $47.69$             & $52.18$             & $49.69$             & \multirow{6}{*}{31B}     & $64.29$             & $41.26$             & $35.66$             & $58.13$             & $25.65$             & $21.90$             \\
                             & Last-turn  & $63.56$             & $60.34$             & $58.07$             & $57.97$             & $51.84$             & $49.99$             &                          & $73.02$             & $49.20$             & $40.35$             & $71.34$             & $45.72$             & $30.82$             \\
                             & Head-25\%  & $\textbf{74.76}$    & $\textbf{71.38}$    & $\textbf{66.38}$    & $57.81$             & $50.85$             & $48.95$             &                          & $75.42$             & $52.74$             & $42.24$             & $77.24$             & $36.89$             & $28.56$             \\
                             & Tail-25\%  & $59.96$             & $57.14$             & $55.09$             & $63.19$             & $55.71$             & $53.47$             &                          & $77.23$             & $42.15$             & $36.35$             & $80.83$             & $38.37$             & $29.84$             \\
                             & Top-25\%   & $62.08$             & $\underline{63.87}$ & $\underline{60.73}$ & $\textbf{67.24}$    & $\textbf{59.29}$    & $\textbf{56.14}$    &                          & $\underline{83.06}$ & $\underline{57.30}$ & $\underline{45.17}$ & $\underline{84.69}$ & $\underline{50.85}$ & $\underline{34.98}$ \\
                             & All-turn  & $\underline{65.88}$ & $62.86$             & $60.26$             & $\underline{66.04}$ & $\underline{58.76}$ & $\underline{55.70}$ &                          & $\textbf{84.98}$    & $\textbf{57.82}$    & $\textbf{45.22}$    & $\textbf{87.47}$    & $\textbf{53.06}$    & $\textbf{35.90}$    \\
    \midrule
    \multirow{6}{*}{27B}     & First-turn & $58.48$             & $62.48$             & $60.92$             & $52.20$             & $55.65$             & $54.08$             & \multirow{6}{*}{\shortstack{26B-\\A4B}} & $61.67$             & $\textbf{46.81}$    & $\textbf{41.81}$    & $58.59$             & $41.95$             & $39.71$             \\
                             & Last-turn  & $56.01$             & $59.14$             & $56.85$             & $51.72$             & $57.64$             & $55.74$             &                          & $53.03$             & $35.25$             & $32.46$             & $44.14$             & $32.93$             & $30.61$             \\
                             & Head-25\%  & $60.88$             & $67.07$             & $64.91$             & $62.10$             & $60.18$             & $58.21$             &                          & $54.92$             & $34.94$             & $32.86$             & $59.72$             & $40.53$             & $37.57$             \\
                             & Tail-25\%  & $70.37$             & $76.63$             & $72.35$             & $57.25$             & $58.24$             & $56.66$             &                          & $58.64$             & $37.84$             & $35.52$             & $60.07$             & $\textbf{46.88}$    & $\textbf{43.40}$    \\
                             & Top-25\%   & $\textbf{75.24}$    & $\textbf{80.07}$    & $\textbf{74.19}$    & $\textbf{71.83}$    & $\textbf{73.12}$    & $\textbf{68.41}$    &                          & $\underline{65.12}$ & $42.17$             & $38.98$             & $\textbf{64.63}$    & $44.43$             & $41.33$             \\
                             & All-turn   & $\underline{74.31}$ & $\underline{79.07}$ & $\underline{74.03}$ & $\underline{70.51}$ & $\underline{72.27}$ & $\underline{67.79}$ &                          & $\textbf{66.22}$    & $\underline{42.68}$ & $\underline{39.17}$ & $\underline{63.37}$ & $\underline{44.82}$ & $\underline{41.47}$ \\
    \bottomrule
  \end{tabular}
  }
  \caption{Comparison of UQ performance of all proposed aggregation strategies for each LLM on the SWE-bench series, where bold and underlined values denote the best and second-best results, respectively.}
  \label{tab:aggregation_strategy_swe_bench}
\end{table*}

\begin{table}[ht]
\centering
\footnotesize
\resizebox{\linewidth}{!}{
\begin{tabular}{ccccccccccc}
\toprule
\multirow{2}{*}{Scale} & \multirow{2}{*}{Strategy} & \multicolumn{3}{c}{InterCode MBPP} & \multicolumn{3}{c}{InterCode Spider} & \multicolumn{3}{c}{InterCode NL2Bash} \\
\cmidrule(lr){3-5} \cmidrule(lr){6-8} \cmidrule(lr){9-11}
                       &                           & AUROC  & AUPRC & AUARC & AUROC  & AUPRC & AUARC & AUROC  & AUPRC & AUARC \\
\midrule
\multicolumn{11}{c}{\textbf{LLM Family: Qwen3.5}} \\
\midrule
\multirow{6}{*}{9B}       & First-turn & $66.24$ & $\underline{56.27}$ & $\underline{50.23}$ & $63.93$ & $74.56$ & $72.85$ & $\textbf{78.21}$ & $37.58$ & $25.86$ \\
                          & Last-turn  & $57.36$ & $37.55$ & $32.98$ & $58.81$ & $68.18$ & $66.97$ & $\underline{76.91}$ & $34.24$ & $24.63$ \\
                          & Head-25\%  & $63.49$ & $54.09$ & $48.71$ & $66.93$ & $77.15$ & $75.11$ & $74.95$ & $37.57$ & $25.41$ \\
                          & Tail-25\%  & $58.94$ & $38.56$ & $34.27$ & $62.93$ & $72.66$ & $70.96$ & $72.33$ & $27.70$ & $21.47$ \\
                          & Top-25\%   & $\underline{66.45}$ & $\textbf{56.78}$ & $\textbf{50.52}$ & $\textbf{81.73}$ & $\textbf{89.91}$ & $\textbf{85.06}$ & $75.60$ & $\underline{44.97}$ & $\underline{27.90}$ \\
                          & All-turn   & $\textbf{67.05}$ & $51.21$ & $47.37$ & $\underline{81.38}$ & $\underline{89.31}$ & $\underline{84.66}$ & $76.69$ & $\textbf{51.05}$ & $\textbf{29.48}$ \\
\midrule
\multirow{6}{*}{4B}       & First-turn & $66.05$ & $50.53$ & $42.96$ & $74.24$ & $81.41$ & $79.20$ & $78.61$ & $49.85$ & $27.40$ \\
                          & Last-turn  & $78.43$ & $51.19$ & $44.27$ & $59.85$ & $69.58$ & $67.72$ & $71.15$ & $23.57$ & $18.15$ \\
                          & Head-25\%  & $62.05$ & $39.68$ & $36.58$ & $69.52$ & $80.94$ & $78.74$ & $81.25$ & $\textbf{56.70}$ & $\underline{28.91}$ \\
                          & Tail-25\%  & $\underline{81.81}$ & $57.34$ & $47.66$ & $70.32$ & $78.43$ & $76.26$ & $71.39$ & $25.81$ & $19.10$ \\
                          & Top-25\%   & $74.52$ & $\underline{58.27}$ & $\underline{47.78}$ & $\textbf{84.71}$ & $\textbf{90.12}$ & $\textbf{86.12}$ & $\underline{86.54}$ & $50.96$ & $28.59$ \\
                          & All-turn   & $\textbf{84.38}$ & $\textbf{71.50}$ & $\textbf{54.47}$ & $\underline{82.09}$ & $\underline{89.94}$ & $\underline{85.88}$ & $\textbf{87.74}$ & $\underline{51.62}$ & $\textbf{29.10}$ \\
\midrule
\multirow{6}{*}{2B}       & First-turn & $64.88$ & $44.44$ & $40.64$ & $\textbf{75.55}$ & $\textbf{65.24}$ & $\textbf{51.65}$ & $67.59$ & $49.03$ & $34.80$ \\
                          & Last-turn  & $62.31$ & $42.72$ & $39.30$ & $53.39$ & $30.64$ & $26.85$ & $65.63$ & $33.51$ & $28.18$ \\
                          & Head-25\%  & $69.96$ & $49.37$ & $44.19$ & $\underline{72.74}$ & $62.09$ & $\underline{49.81}$ & $72.67$ & $49.29$ & $35.40$ \\
                          & Tail-25\%  & $63.11$ & $44.09$ & $40.24$ & $53.02$ & $30.46$ & $26.98$ & $62.36$ & $32.49$ & $26.10$ \\
                          & Top-25\%   & $\textbf{75.98}$ & $\underline{58.18}$ & $\underline{49.53}$ & $69.85$ & $\underline{62.76}$ & $49.40$ & $\textbf{84.78}$ & $\textbf{58.01}$ & $\textbf{40.74}$ \\
                          & All-turn   & $\underline{74.53}$ & $\textbf{62.32}$ & $\textbf{51.44}$ & $64.98$ & $48.85$ & $42.61$ & $\underline{77.74}$ & $\underline{51.66}$ & $\underline{37.51}$ \\
\midrule
\multicolumn{11}{c}{\textbf{LLM Family: Gemma 4}} \\
\midrule
\multirow{6}{*}{8B-E4B}   & First-turn & $71.00$ & $56.34$ & $52.20$ & $73.74$ & $83.32$ & $80.13$ & $62.06$ & $41.87$ & $30.53$ \\
                          & Last-turn  & $70.42$ & $57.65$ & $53.04$ & $68.92$ & $72.98$ & $70.57$ & $71.81$ & $42.87$ & $32.17$ \\
                          & Head-25\%  & $71.00$ & $56.34$ & $52.20$ & $64.45$ & $75.40$ & $73.70$ & $62.23$ & $42.22$ & $30.73$ \\
                          & Tail-25\%  & $70.42$ & $57.65$ & $53.04$ & $70.96$ & $81.90$ & $78.89$ & $\textbf{74.47}$ & $44.55$ & $33.25$ \\
                          & Top-25\%   & $\underline{73.04}$ & $\underline{59.09}$ & $\underline{54.05}$ & $\textbf{79.21}$ & $\textbf{86.31}$ & $\textbf{82.55}$ & $68.62$ & $\underline{48.42}$ & $\underline{33.95}$ \\
                          & All-turn   & $\textbf{79.42}$ & $\textbf{64.58}$ & $\textbf{57.71}$ & $\underline{78.69}$ & $\underline{86.07}$ & $\underline{82.33}$ & $\underline{74.29}$ & $\textbf{52.44}$ & $\textbf{36.04}$ \\
\midrule
\multirow{6}{*}{5B-E2B}   & First-turn & $\textbf{72.51}$ & $\textbf{53.51}$ & $\textbf{44.63}$ & $69.87$ & $60.64$ & $55.97$ & $82.75$ & $\textbf{54.35}$ & $\underline{35.67}$ \\
                          & Last-turn  & $56.58$ & $39.40$ & $35.10$ & $54.23$ & $40.77$ & $37.01$ & $65.12$ & $24.45$ & $20.36$ \\
                          & Head-25\%  & $\underline{70.81}$ & $\underline{52.92}$ & $\underline{44.13}$ & $70.61$ & $60.16$ & $55.64$ & $83.30$ & $51.60$ & $34.87$ \\
                          & Tail-25\%  & $56.10$ & $39.25$ & $34.96$ & $53.20$ & $40.88$ & $37.29$ & $65.31$ & $24.52$ & $20.40$ \\
                          & Top-25\%   & $70.28$ & $49.02$ & $42.07$ & $\textbf{78.65}$ & $\textbf{72.11}$ & $\textbf{63.37}$ & $\textbf{87.76}$ & $\underline{53.45}$ & $\textbf{35.72}$ \\
                          & All-turn   & $67.75$ & $47.68$ & $41.17$ & $\underline{76.03}$ & $\underline{67.77}$ & $\underline{60.78}$ & $\underline{84.79}$ & $47.97$ & $33.73$ \\
\bottomrule
\end{tabular}
}
\caption{Comparison of UQ performance of all proposed aggregation strategies for each LLM on the InterCode series, where bold and underlined values denote the best and second-best results, respectively.}
\label{tab:aggregation_strategy_intercode}
\end{table}
\FloatBarrier

\clearpage
\subsection{Additional Sensitivity Analysis}  \label{app:subsec:sensitivity}
This subsection provides additional sensitivity analyses for the proposed GRUET, including analyses of the number of samples per turn $K$, the embedding similarity threshold $\tau$ for merging action nodes, the sampling temperature $T$, and the percentage $d\%$ used for trajectory-level uncertainty aggregation.

\begin{figure*}[ht]
    \centering
    \begin{subfigure}{\linewidth}
        \centering
        \includegraphics[width=\linewidth]{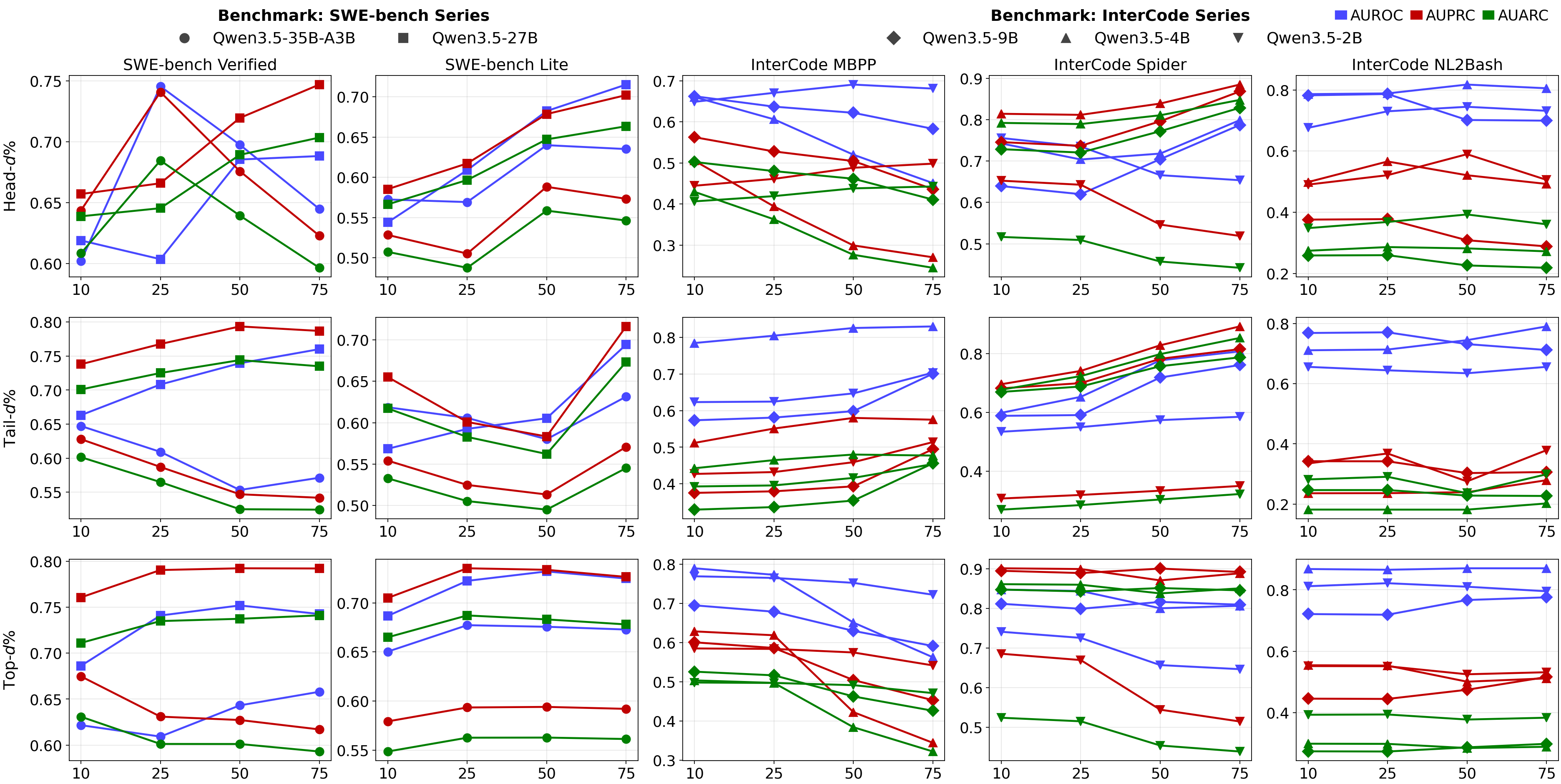}
        \caption{Results for the Qwen3.5 family.}
        \label{fig:sensitivity_d_qwen_app}
    \end{subfigure}

    \vspace{2em}

    \begin{subfigure}{\linewidth}
        \centering
        \includegraphics[width=\linewidth]{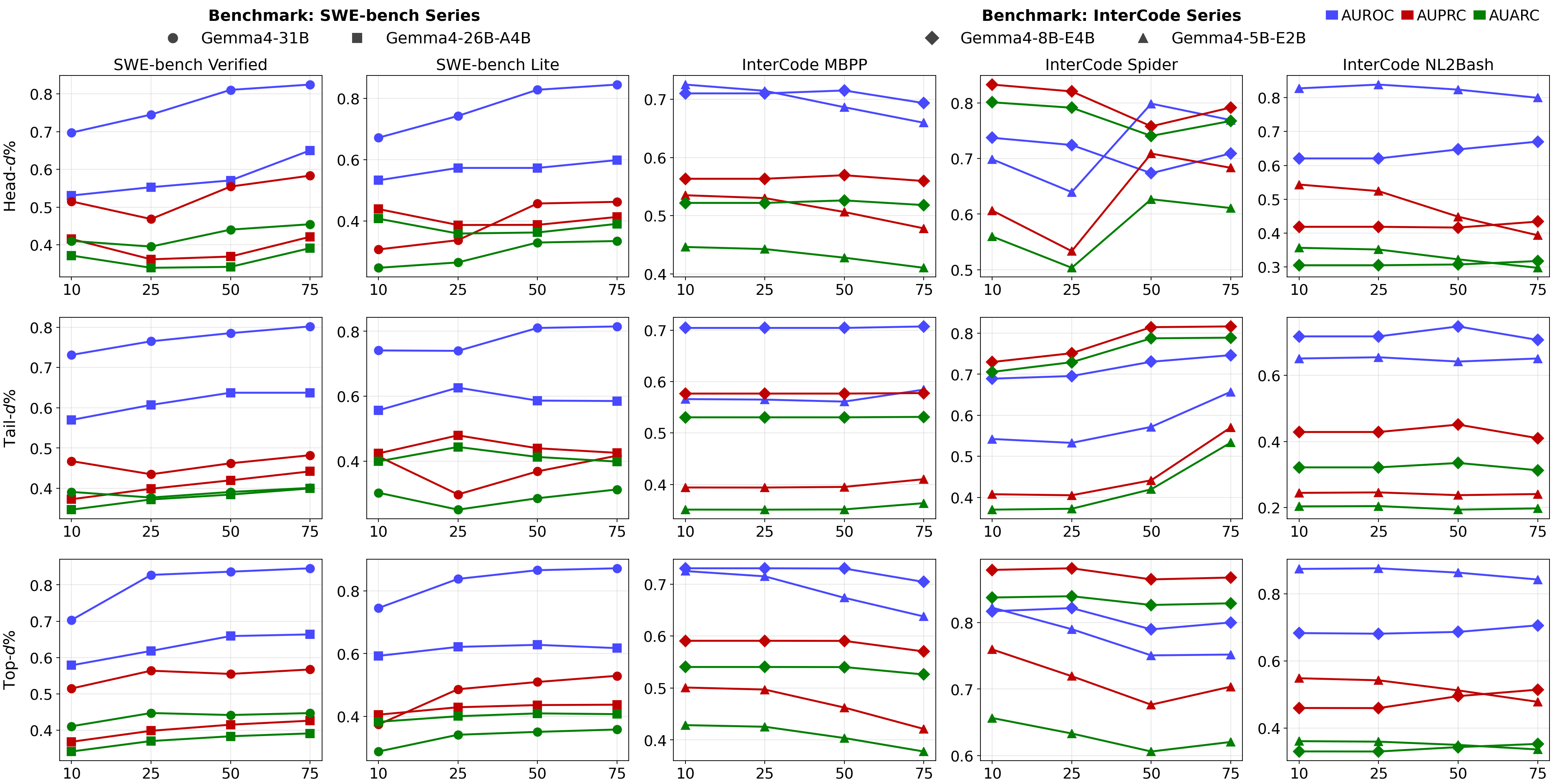}
        \caption{Results for the Gemma 4 family.}
        \label{fig:sensitivity_d_gemma_app}
    \end{subfigure}

    \caption{Sensitivity analyses of the percentage $d\%$ in the aggregation strategy for the Qwen3.5 and Gemma 4 families.}
    \label{fig:sensitivity_d_app}
\end{figure*}

Figures~\ref{fig:sensitivity_d_qwen_app} and~\ref{fig:sensitivity_d_gemma_app} show the sensitivity analyses of the percentage $d\%$ in the aggregation strategy for the Qwen3.5 and Gemma4 families, respectively. It is obvious that all metrics are usually high at $d=25$ for the Head-$d\%$, Tail-$d\%$, and Top-$d\%$ strategies, based on which we recommend $d=25$.

\FloatBarrier

\begin{figure*}[ht]
    \centering
    \includegraphics[width=\linewidth]{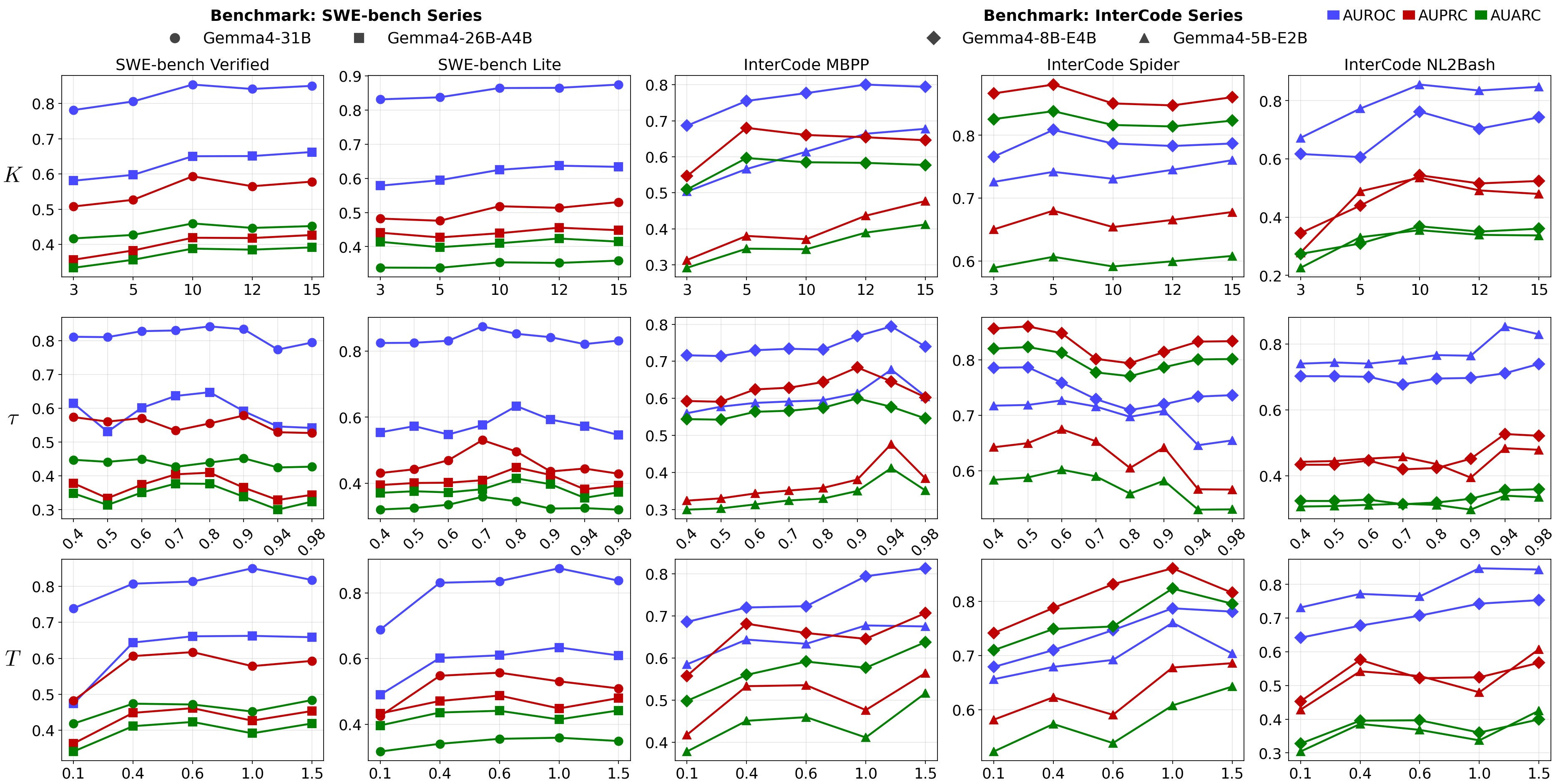}
    \caption{Sensitivity analyses of GRUET with the all-turn strategy for the Gemma 4 family.}
    \label{fig:sensitivity_gemma_app}
\end{figure*}

Figure~\ref{fig:sensitivity_gemma_app} shows the sensitivity analyses of GRUET with the all-turn aggregation for the Gemma 4 family. The first row shows the impact of the number of samples $K$ on UQ performance for the Gemma 4 family. It is observed that all metrics generally increase rapidly as $K$ increases from 3 to 10 and tend to converge when $K$ further increases from 10 to 15. To balance sampling cost and UQ performance, we recommend setting $K=10$. The second row shows the impact of the threshold of embedding similarity $\tau$ on UQ performance for the Gemma 4 family. It is observed that all metrics are relatively insensitive to changes in $\tau$ on all five benchmarks, which demonstrates the robustness of our proposed GRUET with the Gemma 4 family. Given the overall UQ performance across all benchmarks and LLMs, we recommend setting $\tau=0.94$. The third row shows the impact of the sampling temperature $T$ on UQ performance. It is observed that all metrics are relatively low at $T=0.1$, indicating that the proposed GRUET is less effective at low temperatures. This may shed light on the fact that UQ methods requiring multiple sampling are less effective at low temperatures, which aligns with~\citet{lin2024generating}. , which is consistent with the practical setting used in LM-Polygraph\footnote{https://github.com/IINemo/lm-polygraph}~\citep{vashurin2025benchmarking}, a suite of LLM UQ methods. Based on the overall UQ performance across all benchmarks
and LLMs, we recommend setting $T = 1.0$.

\bibliographystyle{apalike}
\bibliography{UReAct}

@article{oh2026uncertainty,
  title={Uncertainty Quantification in LLM Agents: Foundations, Emerging Challenges, and Opportunities},
  author={Oh, Changdae and Park, Seongheon and Kim, To Eun and Li, Jiatong and Li, Wendi and Yeh, Samuel and Du, Xuefeng and Hassani, Hamed and Bogdan, Paul and Song, Dawn and others},
  journal={arXiv preprint arXiv:2602.05073},
  year={2026}
}

@article{ou2026origins,
  title={The Origins of Stochasticity: Comprehensive Investigations on Uncertainty Quantification for Large Language Models},
  author={Ou, Xiang-Jun and Liang, Shuang and Hu, Xin-Yu and Huang, Rong-Hao and Wang, Jing and Zhang, Shao-Qun},
  journal={arXiv preprint arXiv:2606.22792},
  year={2026}
}

@article{shapira2026agents,
  title={Agents of chaos},
  author={Natalie Shapira and Chris Wendler and Avery Yen and Gabriele Sarti and Koyena Pal and Olivia Floody and Adam Belfki and Alex Loftus and Aditya Ratan Jannali and Nikhil Prakash and Jasmine Cui and Giordano Rogers and Jannik Brinkmann and Can Rager and Amir Zur and Michael Ripa and Aruna Sankaranarayanan and David Atkinson and Rohit Gandikota and Jaden Fiotto-Kaufman and EunJeong Hwang and Hadas Orgad and P Sam Sahil and Negev Taglicht and Tomer Shabtay and Atai Ambus and Nitay Alon and Shiri Oron and Ayelet Gordon-Tapiero and Yotam Kaplan and Vered Shwartz and Tamar Rott Shaham and Christoph Riedl and Reuth Mirsky and Maarten Sap and David Manheim and Tomer Ullman and David Bau},
  journal={arXiv preprint arXiv:2602.20021},
  year={2026}
}

@article{zhang2026agenticUQ,
  title={Agentic Uncertainty Quantification},
  author={Zhang, Jiaxin and Choubey, Prafulla Kumar and Huang, Kung-Hsiang and Xiong, Caiming and Wu, Chien-Sheng},
  journal={arXiv preprint arXiv:2601.15703},
  year={2026}
}

@article{zhang2026confidence,
  title={Confidence Estimation for LLMs in Multi-turn Interactions},
  author={Zhang, Caiqi and Yang, Ruihan and Zhu, Xiaochen and Li, Chengzu and Hu, Tiancheng and Dong, Yijiang River and Yang, Deqing and Collier, Nigel},
  journal={arXiv preprint arXiv:2601.02179},
  year={2026}
}

@article{zhang2026agenticConf,
  title={Agentic confidence calibration},
  author={Zhang, Jiaxin and Xiong, Caiming and Wu, Chien-Sheng},
  journal={arXiv preprint arXiv:2601.15778},
  year={2026}
}

@inproceedings{zhao2025uncertainty,
  title={Uncertainty propagation on llm agent},
  author={Zhao, Qiwei and Li, Dong and Liu, Yanchi and Cheng, Wei and Sun, Yiyou and Oishi, Mika and Osaki, Takao and Matsuda, Katsushi and Yao, Huaxiu and Zhao, Chen and others},
  booktitle={Proceedings of the 63rd Annual Meeting of the Association for Computational Linguistics},
  pages={6064--6073},
  year={2025}
}

@article{duan2025uprop,
  title={Uprop: Investigating the uncertainty propagation of llms in multi-step agentic decision-making},
  author={Duan, Jinhao and Diffenderfer, James and Madireddy, Sandeep and Chen, Tianlong and Kailkhura, Bhavya and Xu, Kaidi},
  journal={arXiv preprint arXiv:2506.17419},
  year={2025}
}

@article{shorinwa2025survey,
  title={A survey on uncertainty quantification of large language models: Taxonomy, open research challenges, and future directions},
  author={Shorinwa, Ola and Mei, Zhiting and Lidard, Justin and Ren, Allen Z and Majumdar, Anirudha},
  journal={ACM Computing Surveys},
  volume={58},
  number={3},
  pages={1--38},
  year={2025}
}

@article{fu2025deep,
  title={Deep think with confidence},
  author={Fu, Yichao and Wang, Xuewei and Tian, Yuandong and Zhao, Jiawei},
  journal={arXiv preprint arXiv:2508.15260},
  year={2025}
}

@book{moore2017introduction,
  title     = {Introduction to the Practice of Statistics},
  author    = {Moore, David S. and McCabe, George P. and Craig, Bruce A.},
  year      = {2017},
  publisher = {W. H. Freeman},
}

@article{scarselli2008graph,
  title={The graph neural network model},
  author={Scarselli, Franco and Gori, Marco and Tsoi, Ah Chung and Hagenbuchner, Markus and Monfardini, Gabriele},
  journal={IEEE Transactions on Neural Networks},
  volume={20},
  number={1},
  pages={61--80},
  year={2008},
}

@inproceedings{zhang2024how,
title={How Language Model Hallucinations Can Snowball},
author={Muru Zhang and Ofir Press and William Merrill and Alisa Liu and Noah A. Smith},
booktitle={Proceedings of the 41st International Conference on Machine Learning},
pages = {59670--59684},
year={2024},
}

@inproceedings{gan2025rethinking,
title={Rethinking External Slow-Thinking: From Snowball Errors to Probability of Correct Reasoning},
author={Zeyu Gan and Yun Liao and Yong Liu},
booktitle={Proceedings of the 42nd International Conference on Machine Learning},
pages = {18170--18188},
year={2025},
}

@inproceedings{Yao2023react,
  author       = {Shunyu Yao and
                  Jeffrey Zhao and
                  Dian Yu and
                  Nan Du and
                  Izhak Shafran and
                  Karthik R. Narasimhan and
                  Yuan Cao},
  title        = {ReAct: Synergizing Reasoning and Acting in Language Models},
  booktitle    = {Proceedings of the 11th International Conference on Learning Representations},
  year         = {2023},
}

@inproceedings{liu2025codexembed,
  title     = {CodeXEmbed: A Generalist Embedding Model Family for Multilingual and Multi-task Code Retrieval},
  author    = {Liu, Ye and Meng, Rui and Joty, Shafiq and Savarese, Silvio and Xiong, Caiming and Zhou, Yingbo and Yavuz, Semih},
  booktitle = {Proceedings of the 2nd Conference on Language Modeling},
  year      = {2025},
}

@inproceedings{he2021deberta,
  title={DeBERTa: Decoding-enhanced BERT with Disentangled Attention},
  author={He, Pengcheng and Liu, Xiaodong and Gao, Jianfeng and Chen, Weizhu},
  booktitle={Proceedings of the 9th International Conference on Learning Representations},
  year={2021},
}

@inproceedings{williams2018broad,
  title={A broad-coverage challenge corpus for sentence understanding through inference},
  author={Williams, Adina and Nangia, Nikita and Bowman, Samuel},
  booktitle={Proceedings of the 16th Conference of the North American Chapter of the Association for Computational Linguistics: Human Language Technologies},
  pages={1112--1122},
  year={2018}
}

@inproceedings{yang2024sweagent,
  author       = {John Yang and
                  Carlos E. Jimenez and
                  Alexander Wettig and
                  Kilian Lieret and
                  Shunyu Yao and
                  Karthik Narasimhan and
                  Ofir Press},
  title        = {SWE-agent: Agent-Computer Interfaces Enable Automated Software Engineering},
  pages = {50528--50652},
  booktitle    = {Advances in Neural Information Processing Systems 38},
  year         = {2024}
}

@inproceedings{jimenez2024swebench,
  author       = {Carlos E. Jimenez and
                  John Yang and
                  Alexander Wettig and
                  Shunyu Yao and
                  Kexin Pei and
                  Ofir Press and
                  Karthik R. Narasimhan},
  title        = {SWE-bench: Can Language Models Resolve Real-world Github Issues?},
  booktitle    = {Proceedings of the 12th International Conference on Learning Representations},
  year         = {2024},
}

@inproceedings{yang2023intercode,
 author = {Yang, John and Prabhakar, Akshara and Narasimhan, Karthik and Yao, Shunyu},
 booktitle = {Advances in Neural Information Processing Systems 36},
 pages = {23826--23854},
 title = {InterCode: Standardizing and Benchmarking Interactive Coding with Execution Feedback},
 year = {2023}
}

@article{austin2021program,
  title={Program synthesis with large language models},
  author={Austin, Jacob and Odena, Augustus and Nye, Maxwell and Bosma, Maarten and Michalewski, Henryk and Dohan, David and Jiang, Ellen and Cai, Carrie and Terry, Michael and Le, Quoc and others},
  journal={arXiv preprint arXiv:2108.07732},
  year={2021}
}

@inproceedings{yu2018spider,
  title={Spider: A large-scale human-labeled dataset for complex and cross-domain semantic parsing and text-to-sql task},
  author={Yu, Tao and Zhang, Rui and Yang, Kai and Yasunaga, Michihiro and Wang, Dongxu and Li, Zifan and Ma, James and Li, Irene and Yao, Qingning and Roman, Shanelle and others},
  booktitle={Proceedings of the 22nd Conference on Empirical Methods in Natural Language Processing},
  pages={3911--3921},
  year={2018}
}

@inproceedings{lin2018nl2bash,
  title={Nl2bash: A corpus and semantic parser for natural language interface to the linux operating system},
  author={Lin, Xi Victoria and Wang, Chenglong and Zettlemoyer, Luke and Ernst, Michael D},
  booktitle={Proceedings of the 11th International Conference on Language Resources and Evaluation},
  year={2018}
}

@article{fomicheva2020ppl,
  title={Unsupervised quality estimation for neural machine translation},
  author={Fomicheva, Marina and Sun, Shuo and Yankovskaya, Lisa and Blain, Fr{\'e}d{\'e}ric and Guzm{\'a}n, Francisco and Fishel, Mark and Aletras, Nikolaos and Chaudhary, Vishrav and Specia, Lucia},
  journal={Transactions of the Association for Computational Linguistics},
  volume={8},
  pages={539--555},
  year={2020}
}

@article{farquhar2024detecting,
  title={Detecting hallucinations in large language models using semantic entropy},
  author={Farquhar, Sebastian and Kossen, Jannik and Kuhn, Lorenz and Gal, Yarin},
  journal={Nature},
  volume={630},
  number={8017},
  pages={625--630},
  year={2024},
}

@inproceedings{duan2024sar,
  title={Shifting attention to relevance: Towards the predictive uncertainty quantification of free-form large language models},
  author={Duan, Jinhao and Cheng, Hao and Wang, Shiqi and Zavalny, Alex and Wang, Chenan and Xu, Renjing and Kailkhura, Bhavya and Xu, Kaidi},
  booktitle={Proceedings of the 62nd Annual Meeting of the Association for Computational Linguistics},
  pages={5050--5063},
  year={2024}
}

@article{kadavath2022ptrue,
  title={Language models (mostly) know what they know},
  author={Kadavath, Saurav and Conerly, Tom and Askell, Amanda and Henighan, Tom and Drain, Dawn and Perez, Ethan and Schiefer, Nicholas and Hatfield-Dodds, Zac and DasSarma, Nova and Tran-Johnson, Eli and others},
  journal={arXiv preprint arXiv:2207.05221},
  year={2022}
}

@inproceedings{xiong2024can,
  title={Can LLMs Express Their Uncertainty? An Empirical Evaluation of Confidence Elicitation in LLMs},
  author={Xiong, Miao and Hu, Zhiyuan and Lu, Xinyang and LI, YIFEI and Fu, Jie and He, Junxian and Hooi, Bryan},
  booktitle={Proceedings of the 12th International Conference on Learning Representations},
  year={2024}
}

@article{lin2024generating,
  title={Generating with Confidence: Uncertainty Quantification for Black-box Large Language Models},
  author={Lin, Zhen and Trivedi, Shubhendu and Sun, Jimeng},
  journal={Transactions on Machine Learning Research},
  year={2024}
}

@inproceedings{qiu2024semantic,
  title={Semantic density: Uncertainty quantification for large language models through confidence measurement in semantic space},
  author={Qiu, Xin and Miikkulainen, Risto},
  booktitle={Advances in Neural Information Processing Systems 37},
  pages={134507--134533},
  year={2024}
}

@inproceedings{malinin2021mcse,
  title={Uncertainty Estimation in Autoregressive Structured Prediction},
  author={Malinin, Andrey and Gales, Mark},
  booktitle={Proceedings of the 9th International Conference on Learning Representations},
  year={2021},
}

@article{vashurin2025benchmarking,
  title={Benchmarking uncertainty quantification methods for large language models with lm-polygraph},
  author={Vashurin, Roman and Fadeeva, Ekaterina and Vazhentsev, Artem and Rvanova, Lyudmila and Vasilev, Daniil and Tsvigun, Akim and Petrakov, Sergey and Xing, Rui and Sadallah, Abdelrahman and Grishchenkov, Kirill and others},
  journal={Transactions of the Association for Computational Linguistics},
  volume={13},
  pages={220--248},
  year={2025},
}

@inproceedings{ren2023out,
  title={Out-of-Distribution Detection and Selective Generation for Conditional Language Models},
  author={Ren, Jie and Luo, Jiaming and Zhao, Yao and Krishna, Kundan and Saleh, Mohammad and Lakshminarayanan, Balaji and Liu, Peter J},
  year = {2023},
  booktitle={Proceedings of the 11th International Conference on Learning Representations}
}

@inproceedings{nadeem2009accuracy,
  title={Accuracy-rejection curves (ARCs) for comparing classification methods with a reject option},
  author={Nadeem, Malik Sajjad Ahmed and Zucker, Jean-Daniel and Hanczar, Blaise},
  booktitle={Proceedings of the 3rd International Workshop on Machine Learning in Systems Biology},
  pages={65--81},
  year={2009},
}

@article{chen2025towards,
  title={Towards reasoning era: A survey of long chain-of-thought for reasoning large language models},
  author={Chen, Qiguang and Qin, Libo and Liu, Jinhao and Peng, Dengyun and Guan, Jiannan and Wang, Peng and Hu, Mengkang and Zhou, Yuhang and Gao, Te and Che, Wanxiang},
  journal={arXiv preprint arXiv:2503.09567},
  year={2025}
}

@article{bucur2020epidemic,
  author       = {Doina Bucur and
                  Petter Holme},
  title        = {Beyond ranking nodes: Predicting epidemic outbreak sizes by network
                  centralities},
  journal      = {PLOS Computational Biology},
  volume       = {16},
  number       = {7},
  year         = {2020}
}

\end{document}